%% file: emnlp2023.tex
\pdfoutput=1
\documentclass[11pt]{article}

\usepackage[final]{EMNLP2023}
\usepackage{multirow}
\usepackage{colortbl}
\usepackage{graphics}
\usepackage{booktabs}
\usepackage{url}
\usepackage[ruled,vlined,algo2e]{algorithm2e}
\usepackage{algorithm}
\usepackage{algorithmic}
\let\Algorithm\algorithm
\renewcommand\algorithm[1][]{\Algorithm[#1]\setstretch{1}}

\usepackage{times}
\usepackage{latexsym}
\usepackage{amsmath}
\usepackage[T1]{fontenc}
\usepackage[utf8]{inputenc}
\usepackage{graphicx}

\usepackage{microtype}

\usepackage{inconsolata}
\usepackage{float}

\title{Adaptive Query Rewriting: Aligning Rewriters through Marginal Probability of Conversational Answers}

\author{Tianhua Zhang$^{\heartsuit}$\thanks{$\;\;$Equal contribution.}$\;\,$, Kun Li$^{\heartsuit*}$, Hongyin Luo$^{\diamondsuit}$, \\ \bf Xixin Wu$^{\heartsuit}$, James Glass$^{\diamondsuit}$, Helen Meng$^{\heartsuit}$ \\
$^\heartsuit$The Chinese University of Hong Kong, Hong Kong SAR, China \\
$^\diamondsuit$Massachusetts Institute of Technology, Cambridge MA, USA \\
\texttt{thzhang@link.cuhk.edu.hk, kunli@se.cuhk.edu.hk}
}

\begin{document}
\maketitle
\begin{abstract}
Query rewriting is a crucial technique for passage retrieval in open-domain conversational question answering (CQA). It decontexualizes conversational queries into self-contained questions suitable for \textit{off-the-shelf} retrievers. Existing methods attempt to incorporate retriever's preference during the training of rewriting models. However, these approaches typically rely on extensive annotations such as in-domain rewrites and/or relevant passage labels, limiting the models' generalization and adaptation capabilities. In this paper, we introduce AdaQR (\textbf{Ada}ptive \textbf{Q}uery \textbf{R}ewriting), a framework for training query rewriting models with limited rewrite annotations from seed datasets and completely no passage label. Our approach begins by fine-tuning compact large language models using only  \textasciitilde$10\%$ of rewrite annotations from the seed dataset training split. The 
models are then utilized to generate rewrite candidates for each query instance. A novel approach is then proposed to assess retriever's preference for these candidates by the probability of answers conditioned on the conversational query by marginalizing the Top-$K$ passages. This serves as the reward for optimizing the rewriter further using Direct Preference Optimization (DPO), a process free of rewrite and retrieval annotations. Experimental results on four open-domain CQA datasets demonstrate that AdaQR not only enhances the in-domain capabilities of the rewriter with limited annotation requirement, but also adapts effectively to out-of-domain datasets.
\end{abstract}

\section{Introduction}
Passage retrieval in open-domain conversational question answering (CQA) have gained significant prominence in recent years \cite{anantha-etal-2021-open}. Unlike standard retrieval with single-turn queries \cite{kwiatkowski-etal-2019-natural}, it poses unique challenges in resolving conversational dependencies like omission, ambiguity, and coreference resolution \cite{Qu_2020, adlakha2022topiocqa}. Many existing methods \cite{Yu_2021, lin2021contextualized,li-etal-2022-grounded} address these challenges by training specialized retrievers. However, re-training retrievers for conversational search can be costly and may not fully leverage the benefits of \textit{off-the-shelf} single-turn retrievers \cite{wu2022conqrr}.

A prevalent approach for overcoming this challenge involves \textit{query rewriting} (QR)
\cite{elgohary-etal-2019-unpack, vakulenko2020question, yu2020fewshot}. In this method, conversational queries are decontextualized into self-contained, standalone queries, which are then processed by \textit{off-the-shelf} retrievers to find relevant information.
Earlier studies \cite{elgohary-etal-2019-unpack, anantha-etal-2021-open} focused on fine-tuning language models to reformulate human rewrites. However,
\citet{ye-etal-2023-enhancing} noted that human annotations may only resolve ambiguity while overlooking informative context within conversations. They suggested using language language models (LLMs) for rewrite generation. Recent research \cite{wu2022conqrr,mo2024convgqr, yoon2024ask} underscores the significance of incorporating retrieval signals during rewriter training to enhance downstream retrieval performance. \citet{yoon2024ask} aligned fine-tuned models with retriever's feedback on the ranking of gold passages using Direct Preference Optimization (DPO).

Nevertheless, these approaches often necessitate substantial amounts of rewrite and/or passage labels for supervision, yet resources are scarce and expensive for collection \cite{Yu_2021}.
Moreover, they mainly optimize QR systems for in-domain performance, i.e., training labels are from the validated datasets, while the adaptation ability and out-of-domain performance are under-explored. 
Therefore, this paper centers on: (1) \textit{effectively and efficiently training QR models with limited annotation requirements and}, (2) \textit{examining their capacity for adaptation with preference alignment under weak supervision.}

Following the paradigm for aligning LLMs which takes supervised fine-tuning (SFT) and preference optimization sequentially \cite{Ouyang2022TrainingLM}, optimization towards retrievers' preferences can be applied to the models that have undergone supervised fine-tuning for query rewriting. Aligning with retrievers' preferences can further tune the rewriter to reformulate queries with better recall of relevant passages \cite{yoon2024ask}. Importantly, we also find it to be capable of adapting rewriters to out-of-domain CQA scenarios (§\ref{Main Results}). However, a core issue is how the retrievers' preferences should be modeled. 
While \citet{yoon2024ask} uses the ranking of gold passage as the retrievers' preferences, we aim to explore the extent to which the use of labeled data can be reduced. We argue that the corresponding answers within the conversation can be employed to formulate the retriever's preferences. Moreover, conversation answers are more readily accessible than the gold passages, as they naturally happen when the CQA data is synthesized.

\begin{figure*}[ht]
\centering 
\includegraphics[width=0.8\textwidth]{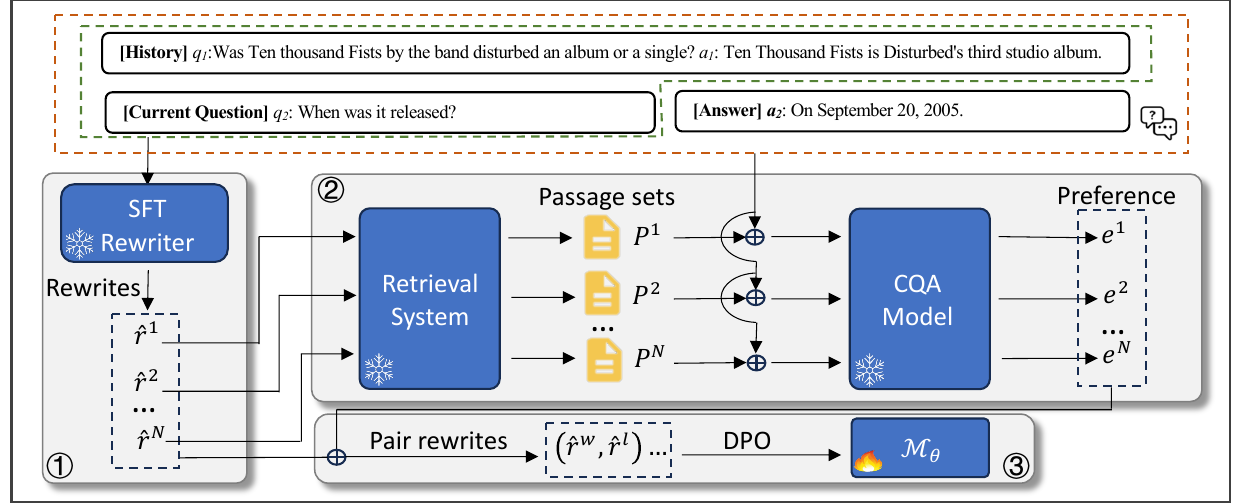}
\caption{Illustration of AdaQR which applies preference optimization to the rewriter $\mathcal{M}_{\theta}$.}
\label{overview_figure}
\vspace{-0.5\baselineskip}
\end{figure*}
We propose a novel preference optimization approach, AdaQR, for query rewriters, aiming to optimize rewriters to 
% generate rewrites catering 
cater for retrievers, by utilizing conversation answers to model retrievers' preferences. Specifically, we first let the SFT rewriter to generate several rewrites, which are then used as the queries to retrieve a set of passages by a target retriever. Subsequently, we calculate the conditional probability of the answer for each retrieved passage and the conversation, and obtain the marginal probability of the answer by marginalizing the passages set. The marginal probability of the answer serves as the reward quantifying the retrievers' preferences over rewrites. Finally, we pair these rewrites based on their reward for optimizing the SFT rewriter with DPO.

We examine the in-domain performance of AdaQR where the training data for SFT and preference optimization all comes from the validated datasets. Empirical results show that  AdaQR greatly improves the quality of rewrites generated by the rewriter, compared with the SFT-only counterpart, leading to comparable or even  better performance over existing SOTA QR methods. More importantly,  the out-of-domain evaluation, where the preference optimization is applied to a out-of-domain SFT rewriter, also observes the same performance gain, justifying the ability of AdaQR to adapt the rewriter to the target domains.

% The advantage of AdaQR can be abbreviated as (1) \textbf{Efficiency and flexibility}: AdaQR models retrievers' preferences by making use of the answer within conversations, allowing training query rewriters for various conversational question answering tasks even without passage labels. (2) \textbf{Enhancing both in- and out-of-domain performance}: AdaQR can not only amplify rewriters in-domain capability, but also adapt them to out-of-domain conversational question answering tasks.
The key contributions of this work are: (1) We propose AdaQR, a preference optimization approach for enhancing query rewriter. (2) AdaQR models retrievers' preferences by leveraging the answer within conversations, allowing training query rewriters for various conversational question answering tasks even without passage labels. (3) Experiments show AdaQR can not only amplify rewriters in-domain capability, but also adapt them to out-of-domain conversational question answering tasks.

\section{Methodology}

\subsection{Task Formulation}
We focus on the \textit{query rewriting} task for conversational passage retrieval using \textit{off-the-shelf} retrievers. Given the conversation history $H_{<t}=\{q_i,a_i\}_{i=1}^{t-1}$, where $t$ denotes the current turn number, a query rewriting model $\mathcal{M_\theta}$ is trained to transform the the current question $q_t$ into a standalone, self-contained query $\hat{r}_t$. We omit the subscript $t$ in subsequent description for simplicity. The retriever $\mathcal{R}$, which remains unchanged, takes $\hat{r}$ as input to search for relevant passages from the corpus $P$. In the complete training process of QR, we only need limited rewrite labels $\{r\}$ from two seed datasets, with no passage annotations required.

\subsection{Overview}

We propose AdaQR to build query rewriters applicable to various conversational question answering scenarios, through preference optimization involving no in-domain rewrite or passage labels. AdaQR uses the probability of answers as the reward quantifying the retriever's preferences over the rewrites, to further optimize 
% a rewriter already tuned on out-of-domain data and adapt it to a target dataset. 
an already tuned rewriter on out-of-domain data and adapt it to a target dataset. 
As shown in Fig. \ref{overview_figure},  AdaQR, warm-started with a SFT query rewriter fine-tuned with a limited number of (in- or out-of-domain) labeled data beforehand (§\ref{Supervised Fine-Tuning}), operates in the pipeline: (1) Sample rewrites from the SFT rewriter, which are then used as search queries for retrieving passages; (2) Derive the marginal probability of the answer as reward based on the conversation and retrieved passages (§\ref{Reward Collection}); (3) Construct preference pairs using the reward and tune the rewriter with Direct Preference Optimization \cite{rafailov2024direct} (§\ref{Preference Optimization}).
\subsection{Supervised Fine-Tuning}
\label{Supervised Fine-Tuning}
\textbf{Limited Rewrite Labels} To empower models with basic query rewriting capability, we need a limited number of rewrite labels for supervised fine-tuning. We separately curate labels (\textasciitilde10\% of the training set) from two seed datasets, QReCC \cite{anantha-etal-2021-open} and TopiOCQA \cite{adlakha2022topiocqa} for comprehensive adaptation performance analysis of AdaQR. 
We derive \texttt{QReCC-SFT} with $3,850$ rewrite labels generated by ChatGPT (\texttt{gpt-3.5-turbo}) under few-shot learning setting\footnote{\citet{ye-etal-2023-enhancing} provides rewrite labels in both few-shot learning (FSL) setting and advanced editor setting. 
We use the labels generated in the initial FSL setting.} 
from previous work \cite{ye-etal-2023-enhancing}. We derive \texttt{TopiOCQA-SFT} 
% $\mathcal{M_{T}}$ 
with $4,278$ training instances, with rewrite labels generated by \texttt{gpt-4}. The same instruction and in-context learning examples used in QReCC from \citet{ye-etal-2023-enhancing} are followed.
The complete prompt is detailed in Appendix Table \ref{tab: rewrite-label-prompt}. Note that no passage label is required in this stage. The resulting two fine-tuning models $\mathcal{M}_{SFT}$,
are subsequently adapted and tested on additional datasets.  
% QReCC: 3859 vs 51928 (total) vs 39677 (valid train)
% TopiOCQA: 4278 vs 45450 (total) vs 41941? (valid answer) vs 38862 (valid train)

\noindent\textbf{Training Objective} We use the curated labeled data to fine-tune a language model, equipping it with basic query rewriting ability. Given the conversation history $H$ and the current query $q$, the LM $\mathcal{M_\theta}$ is trained to predict the rewrite $r$ by minimizing the negative log-likelihood as
\begin{equation}
\mathcal{L}_{SFT} = -\log p_{\mathcal{M_\theta}}(r|H, q)
\end{equation}

It should be noted that the models trained in the above way might have insufficient rewriting abilities, especially for out-of-domain conversational query (will be shown in §\ref{Main Results}). Next, we apply AdaQR to these SFT rewriters to enhance their capabilities for both in- and out-of-domain scenarios.

\subsection{Reward Collection}
\label{Reward Collection}
\noindent \textbf{Rewrite Sampling} To obtain training data for preference optimization, we use the fine-tuned models, $\mathcal{M}_{SFT}$, 
to sample three rewrite candidates $\{\hat{r}^i\}_{i=1}^3$
for each conversational query $q$ with the temperature $T=1$.
This strategy produces reasonable rewrite candidates for preference optimization, bypassing the need of expensive human labeling or large language model prompting adopted by previous work \cite{yoon2024ask}.

\noindent \textbf{Reward Calculation} We propose a novel reward calculation method that relies solely on conversation turns, eliminating the need for passage labels. 
Motivated by \citet{lewis2021retrievalaugmented}, 
we use weak supervision with the \textbf{marginal probability of \textit{answers}} as the preference feedback, treating the retrieved passages as a latent variable. Concretely, for each rewrite candidate $\hat{r}^i$, we retrieve top-$K$\footnote{$K=5$ except in §\ref{sec: analysis-topk} where we evaluate the effect of $K$.} passages $P^i = \{p_{k}^i\}_{k=1}^{K}$ with the retriever $\mathcal{R}$. A pre-trained large language model $\mathcal{A}$ then calculates the 
log probability $\mathcal{S}_k$
of generating the target answer $a$ conditioned on each retrieved passage $p_{k}^i$ and the original question $q$ concatenated after the conversation history $H$: 
\begin{equation}
% \mathcal{L}_{CQA}^k=-\log p_{\mathcal{A}}(a_t|H_{<t}, q_t,p^i_{k}) 
\mathcal{S}_k=\log p_{\mathcal{A}}(a|H, q,p^i_{k}) 
\end{equation}
We select a pre-trained model due to the conjecture that LLMs have inherent capabilities established during pretraining, while later fine-tuning or alignment may affect the distribution of logits, resulting in alignment tax or capability misalignment \cite{huang2023survey, lin2024mitigating, gekhman2024does}.
To control for the influence of the rewrite in the probability calculation, we use the original conversation as input question rather than the rewritten queries. This ensures the grounding passage $p_{k}^i$ is solely responsible for the contribution to the score $\mathcal{S}_k$.
By marginalizing the top $K$ passages, we calculate the marginal probability of answer $e^{i}$ as the retrievers' preference for rewrite candidate $\hat{r}^{i}$:
\begin{equation}
% e^{i} = \sum_{k=1}^{K} \mathcal{P}_{\mathcal{R}}(p_{k}^i|\hat{r}^i) \mathcal{L}_{CQA}^k
e^{i} = \Sigma_{k=1}^{K} \mathcal{P}_{\mathcal{R}}(p_{k}^i|\hat{r}^i) \mathcal{S}_k
\label{eq:reward}
\end{equation}
% \[e_t^{i} = \sum_{p_{tk}^i\in P_t^i} \mathcal{P}_{Ret}(p_{tk}^i|\hat{r}_t^i) \mathcal{L}(a_t|p_{tk}^i, H_{<t},q_t)\]
where $\mathcal{P}_{\mathcal{R}}(p_{k}^i|\hat{r}^i)$ denotes the distribution over passages obtained by applying a softmax function to their retrieval scores. Intuitively, a more effective rewrite used as a search query improves the chances of recalling potentially relevant passages. A passage with heightened relevance further enhances the likelihood of generating the answer. Consequently, this rewrite leads to a better reward with a higher marginal probability. 

\subsection{Preference Optimization}
\label{Preference Optimization}
Our goal is to align the SFT rewrite model $\mathcal{M}_{\theta}$ to generate rewrites preferred by the target retriever, quantified by the marginal answer probabilities $e$ as in Eq. \ref{eq:reward}. To this end, we apply Direct Preference Optimization \cite{rafailov2024direct} with $e$ as reward to tune $\mathcal{M}_{\theta}$. 
\input{main-qrecc-topiocqa}

\noindent\textbf{Preference Pairs Construction} For each conversation example, we construct pairwise preference data $\{(H, q, \hat{r}^w, \hat{r}^l )\}$ by selecting pairs of rewrites $(\hat{r}^w, \hat{r}^l )$ from $\{\hat{r}^i\}_{i=1}^3$ such that $e^w-e^l>\delta$, where $\delta>0$ is a hyperparameter. Due to the characteristics of $e^i$, this constraint ensures that the preferred rewrite $\hat{r}^w$ will lead to useful passages more likely than the dispreferred one $\hat{r}^l$. Unlike conventional ones, our preference data is developed without any human annotations, by using the automatic measurement of retriever preferences in §\ref{Reward Collection}. 

\noindent\textbf{Training Objective} Using the pairwise preference data, we tune the model $\mathcal{M}_{\theta}$ with DPO. The training objective is to minimize
\begin{equation}
\begin{split}
L_{D P O}=-\log \sigma(\beta \log \frac{\mathcal{M}_{\theta}\left(\hat{r}^{w} \mid q, H\right)}{\mathcal{M}_{SFT}\left(\hat{r}^{w} \mid q, H\right)} \\
-\beta \log \frac{\mathcal{M}_{\theta}\left(\hat{r}^{l} \mid q, H\right)}{\mathcal{M}_{SFT}\left(\hat{r}^{l} \mid q, H\right)})
\end{split}
\end{equation}
where $\mathcal{M}_{SFT}$ is the reference model from which $\mathcal{M}_{\theta}$ is initialized, $\sigma$ is the sigmoid function, and $\beta$ is a hyperparameter. With this objective, the model is optimized to maximize the contrast between preferred and dispreferred rewrites. It thus is encouraged to generate rewrites with higher marginal probabilities of the answers, which are more likely to lead to the useful passages. 
% See Algorithm \ref{algorithm} in Appendix for details.
See the complete algorithm of AdaQR in Appendix Algorithm \ref{algorithm}.

\section{Experiments}
\input{main-d2d-md2d}

\noindent \textbf{Datasets}
We evaluate AdaQR for conversational retrieval task on four conversational question answering (CQA) benchmarks: QReCC \cite{anantha-etal-2021-open}, TopiOCQA \cite{adlakha2022topiocqa}, Doc2Dial \cite{feng-etal-2020-doc2dial} and MultiDoc2Dial \cite{feng-etal-2021-multidoc2dial}.
The answers in QReCC exhibit relatively  large word-level overlap to supporting passages while TopiOCQA uses free-form responses as answers (See Analysis \ref{sec:analysis-weak}).  TopiOCQA and MultiDoc2Dial involve \textit{topic-shift} with turns in a conversation grounded on multiple documents.

\noindent \textbf{Retrieval Systems} We investigate the performance of AdaQR using both sparse and dense retrievers. BM25 serves as the sparse retriever for all datasets. For dense retriever, we employ ANCE \cite{xiong2020approximate} trained on MS-MARCO \cite{bajaj2018ms} passage retrieval tasks for all datasets except Doc2Dial, 
to align with previous studies \cite{jang2024itercqr,yoon2024ask}. We use E5-unsupervised \cite{wang2024text} for Doc2Dial
following \citet{liu2024chatqa}.
Notably, we refrain from additional training of the retrievers for our specific task. More details are listed in Appendix \ref{sec:appendix-retrieval}.

\noindent \textbf{Evaluation Metrics} 
We assess retrieval performance using several metrics: Mean Reciprocal Rank (\textbf{MRR}) calculates the average rank of gold passages. Normalized Discounted Cumulative@3 (\textbf{NDCG}) 
evaluates the top-3 retrieval results by considering both relevance and rank. \textbf{Recall@$k$} reflects whether the retriever successfully identifies the gold passages within top-$k$ results.

\noindent \textbf{Implementation} We fine-tuned two SFT models with two seed datasets: \texttt{QReCC-SFT} and \texttt{TopiOCQA-SFT}. Each SFT model underwent further training with DPO, using retriever feedback collected in §\ref{Reward Collection} across all four datasets respectively. Our backbone model for all configurations is Mistral-7B \cite{jiang2023mistral}. To ascertain the generalization ability across different LMs, we further assessed the performance of Gemma-7B \cite{gemmateam2024gemma} and Llama2-7B \cite{touvron2023llama} for all benchmarks with sparse retrieval in Appendix Table \ref{tab:gemma-llama}. 
We used the pretrained \texttt{Mistral-7B-v0.1} as the CQA model for reward calculation. 
See training details in 
Appendix \ref{sec:appendix-training}. 

\noindent \textbf{Baselines} 
(1) \textbf{T5QR} \cite{lin2020conversational} fine-tunes T5-base \cite{JMLR:v21:20-074} to mimic human rewrites.
(2) \textbf{CONQRR} \cite{wu2022conqrr} optimizes query rewriters using reinforcement learning, with the ranking of passages having maximum token overlap to answers as weak supervision. 
(3) \textbf{ConvGQR} \cite{mo2024convgqr} trains query rewriting and expansion models with Mean Squared Error between embeddings of query and relevant passage as an auxiliary loss. 
(4) \textbf{IterCQR} \cite{jang2024itercqr} iteratively trains the query rewriter with cosine similarity between gold passages and reformulated queries by ChatGPT as IR signal.
(5) \textbf{LLM IQR} \cite{ye-etal-2023-enhancing} introduces ``rewrite-then-edit'' to prompt ChatGPT first generates rewrites and then edits them according to pre-defined criteria. 
(6) \textbf{RETPO} \cite{yoon2024ask} prompts \texttt{gpt-4} to generate multiple rewrites and collect gold passage ranking as retrieval feedback upon all training data of QReCC and TopioCQA. \textbf{RETPO} fine-tunes Llama2-7b \cite{touvron2023llama} to replicate the rewrite with the best retrieval preference, termed as (7) \textbf{Llama2 Distill}, and then aligned it with retrieval preference using DPO. 
(8) \textbf{GPT4 Prompting} \cite{yoon2024ask} generates rewrites for test questions with \texttt{gpt-4}.
(9) We implement \textbf{Gold-Label} under our setting, with the ranking of the gold passages as the retrieval preference. This serves as an upper-bound for the performance of our weakly supervised AdaQR.

\section{Main Results}
\label{Main Results}

\input{analysis-weak}

Tab.\ref{tab: main-qrecc-topiocqa} and \ref{tab: main-d2d-md2d} show the main result of AdaQR across 4 benchmarks and 2 retrievers, alongside the comparisons with baseline approaches.

\noindent\textbf{Preference optimization brings improvement over SFT.} AdaQR (\texttt{Ours}) shows consistent and evident improvement over its \texttt{SFT}-only counterpart across all the combinations of datasets and retrievers, indicating the effectiveness of preference optimization in further enhancing rewriters' abilities. 

\noindent\textbf{AdaQR improves in- and out-of-domain performance.} For in-domain scenarios, on QReCC, our approach, \texttt{QReCC-SFT+Ours}, outperforms all baselines; while on TopiOCQA, \texttt{TopiOCQA-SFT+Ours} surpasses baselines other than \texttt{RETPO} and its ablated variant \texttt{Llama2 Distill} under BM25 retriever, in terms of average performance. These baselines all necessitate passage labels, especially \texttt{RETPO} and \texttt{Llama2 Distill} which involve extensive use of both passage labels and rewrite labels from the combination of above two datasets. 
For out-of-domain scenarios across four datasets, \texttt{Ours} that began with a heterogeneous-seed SFT and then underwent preference optimization on target datasets (e.g., \texttt{QReCC SFT+Ours} on TopiOCQA, \texttt{TopiOCQA SFT+Ours} on QReCC), still exceeds most of baselines. In most cases, the heterogeneous-seed \texttt{SFT} lags behinds the baselines, but gets close to or surpasses them after preference optimization (\texttt{+Ours}).
Together, our approach can not only amplify rewriters' in-domain capabilities but also successfully adapt them to out-of-domain tasks, even in the absence of passage labels.

\noindent\textbf{Conversation answer is as effective as passage labels for preference optimization.} Both our approach and its variant with \texttt{Gold-Label}, are implemented in the same paradigm but with different types of reward (marginal probability of answer vs. ranking of gold passages). The two approaches have comparable performances, and even sometimes \texttt{Ours} outperforms \texttt{Gold-Label}, demonstrating the effective role of the marginal probability-based reward for modeling the retrievers' preference, while such reward is more accessible. These makes AdaQR a quite cost-effective method for adapting rewriters to various CQA tasks. 

\begin{figure}[ht]
\centering
\includegraphics[width=0.5\textwidth]{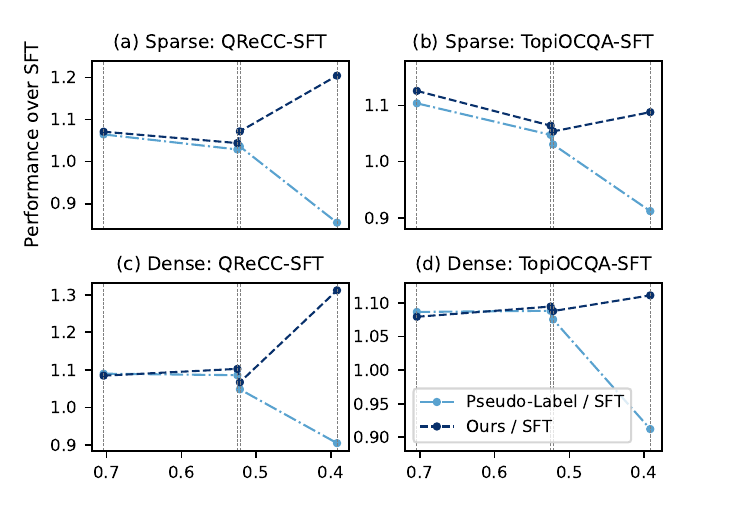}
\caption{Average performance of \texttt{Pseudo-Label} and \texttt{Ours} over \texttt{SFT} as the F1-score (x-axis) between the answers and gold passages declines. Scores $>1$ denote improvement over \texttt{SFT}. The four vertical lines correspond to the F1-scores of QReCC ($0.704$), Doc2Dial ($0.525$), MultiDoc2Dial ($0.522$) and TopiOCQA ($0.392$).
}
% \vspace{-0.5\baselineskip}
\label{fig: analysis-weak}
\end{figure}

\noindent\textbf{AdaQR has applicability to both sparse and dense retrievers.} Positive effect of our approach can be seen for both sparse (BM25) and dense (E5, ANCE) retrievers, which verifies the general applicability of our approach to various retrievers. On the other hand, note that the ANCE retriever has better performance than the BM25 for TopiOCQA, as exemplified by \texttt{SFT}'s higher retrieval metrics under ANCE over under BM25. We also observe that for TopiOCQA, the improvement of \texttt{Ours} over \texttt{SFT} under ANCE is greater than under BM25. Therefore, it is reasonable to speculate that the benefit brought by our approach would get more pronounced with better retrievers.

Similar patterns also manifest in the cases of using Gemma and Llama2 as the base models (see Tab. \ref{tab:gemma-llama}). This points to the general effectiveness of AdaQR with different open-source LMs.

\begin{figure}[ht]
% \centering 
\includegraphics[width=0.49\textwidth]{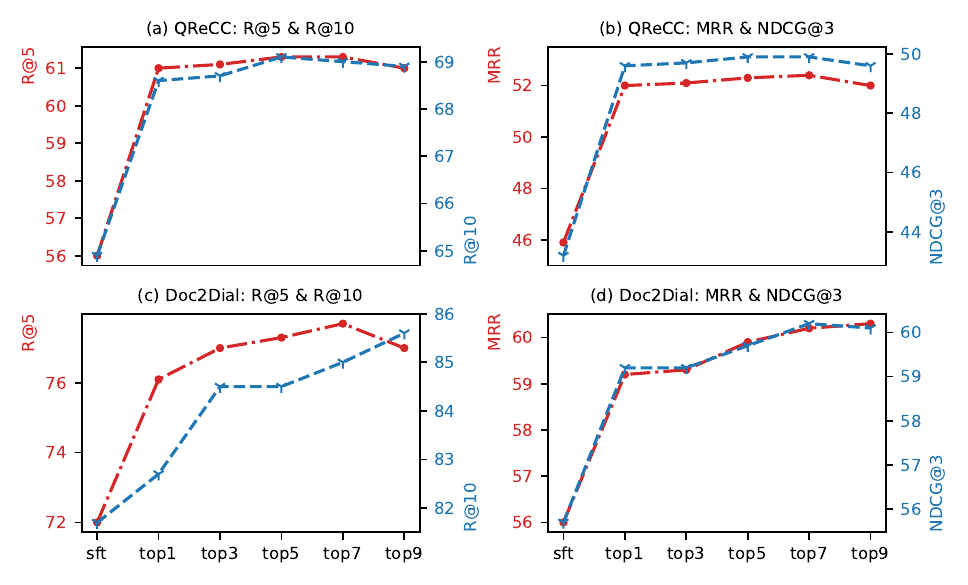}
\caption{Retrieval performance with varying top-$K$ values 
($k=1,3,5,7,9$) 
in reward calculation using \texttt{QReCC-SFT}. 
See detailed results in Appendix Table \ref{tab: analysis-topk}. We further analyze two passage organization types (concatenation and marginalization) in Appendix \ref{sec: appendix-concate}.
% Comprehensive performance metrics are available in Appendix Table \ref{tab: analysis-topk}. 
% Triangles denote the performance of models where DPO training using the answer probability conditioned on concatenated top-$5$ passages as reward.
}
\label{fig: analysis-topk}
\vspace{-0.7\baselineskip}
\end{figure}

\section{Analysis}

\subsection{Comparison of Weakly Supervised Approaches}
\label{sec:analysis-weak}
We compare our reward calculation approach (\texttt{Ours}) to a word-level based weak supervision method (\texttt{Pseudo-Label}) under the same setting, as shown in Table \ref{tab:pseudo-vs-ours}. 
Both approaches eliminate the need for in-domain labels. Our method derives retrieval feedback through assessing the probability of the target answer by marginalizing the top-$K$ passages. On the contrary, \texttt{Pseudo-Label} uses the ranking of pseudo-relevant passages that have the maximum F1-score to the answer as the retriever's preferences, following \citet{wu2022conqrr}. 

\textbf{AdaQR surpasses \texttt{Psuedo-Label} on TopiOCQA, Doc2Dial and MultiDoc2Dial across all settings in terms of average performance.} The relatively good performance of \texttt{Pseudo-Label} on QReCC is attributed to the dataset's characteristics, where the answers exhibit a high level of overlap with sentences in the supporting passages. Consequently, this straightforward word-level based weak supervision can readily identify relevant passage, with 82\% of gold passages detected. However, \textbf{\texttt{Pseudo-Label} is notably susceptible to the influence of the word-level overlap requirement.} We visualize the performance over \texttt{SFT} across four datasets with decreasing F1-scores in
% the F1-score distribution of the test examples across four datasets in 
Figure \ref{fig: analysis-weak}. The retrieval performance of \texttt{Pseudo-Label} drops significantly as the F1-score decreases. Notably, when assessed on TopiOCQA, which features free-form responses as answers, this approach even negatively impacts the results, resulting in performance inferior to the SFT-only version, i.e., \texttt{Pseudo-Label}/\texttt{SFT} $<1$. In contrast, AdaQR measures the retrievers' preferences from semantic-level, \textbf{demonstrating greater robustness and stability, providing consistent performance improvement across all settings.} 
% In contrast, \textbf{our approach demonstrates greater robustness and stability, providing consistent performance improvement across all settings.} 

Moreover, as analyzed in Appendix \ref{sec:appendix-topic-shift} and Figure \ref{fig: analysis-topic-shift}, \textbf{AdaQR shows notable enhancement in addressing challenging queries with \textit{topic-shift}} over \texttt{SFT}, achieving performance ratio comparable to the \texttt{Gold-Label} counterpart. However, \texttt{Pseudo-Label}'s heavy reliance on word-level overlap hampers its effectiveness.

\begin{figure}[ht]
\centering
\includegraphics[width=0.48\textwidth]{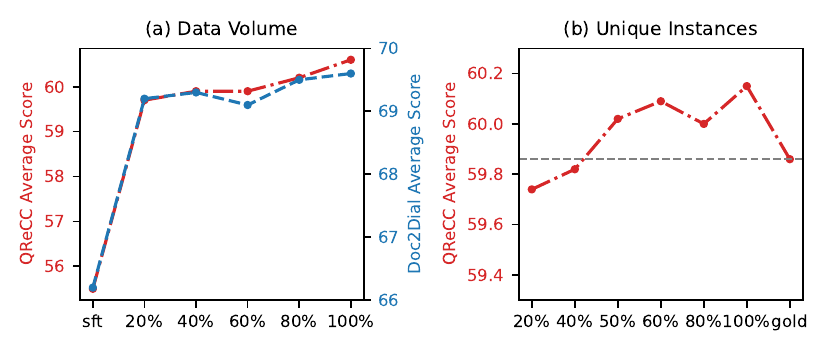}
\caption{Average retrieval performance with varying number of training data during preference optimization under \texttt{QReCC-SFT} setting.}
\vspace{-0.5\baselineskip}
% \includegraphics[width=0.8\textwidth]{emnlp2023-latex/figures/analysis-data-volume-avg.pdf}
% \caption{Average retrieval performance with varying number of training data during preference optimization under \texttt{QReCC-SFT} setting. (a) We first collect \textbf{ALL} rewrite pairs with reward difference larger than a threshold ($100\%$) and then gradually reduce the data volume. Comprehensive performance metrics are available in Appendix Table \ref{tab: appendix-data-volume}. (b) To ascertain the benefit of extra unlabeled data enabled by our weak supervision, we collect \textbf{ONE} pair from each example exhibiting the largest reward difference for both \texttt{Gold-Label} and \texttt{Ours} ($100\%$), and then gradually decrease the volume of data ($80\%$, $60\%$, $40\%$ and $20\%$).}
\label{fig: analysis-data-volume}
\end{figure}
\subsection{Effect of $K$ Values in Reward Calculation}
\label{sec: analysis-topk}
Figure \ref{fig: analysis-topk} depicts the retrieval performance across varying numbers of top retrieved passages ($K=1,3,5,7,9$) for reward calculation in §\ref{Reward Collection}. Detailed performance metrics are presented in Table \ref{tab: analysis-topk} of Appendix. 
% \ref{sec:appendix-topk}. 
Direct preference optimization across all candidate values of $K$ significantly improves the retrieval performance over the \texttt{SFT} verison, reflecting the efficacy of our proposed reward calculation methodology.
Relying solely on the top-$1$ retrieved passage may yield sub-optimal results.
% given that 
For instance, only 36.4\% of training instances rank the gold passage first with BM25 on QReCC. For Doc2Dial, increasing the value of $K$ tends to enhance the overall performance. \textbf{AdaQR demonstrates robustness against potential irrelevant information when more passages are involved by enlarging $K$}, and
% \textbf{Our approach, AdaQR, demonstrates robustness against variations in $K$} and 
effectively reflects retrievers' preferences without requiring in-domain passages or rewrite labels.
% is a robust approach against possible noise to estimate the retrieval feedback without requiring in-domain passage or rewrite labels.  
We opt for $K=5$ across all settings to strike a balance between effectiveness and efficiency, avoiding manual bias towards the best configuration.

% /mnt/file-201-project-disk-m/project/dialogue/AI4Future/research-projects/conv-rewrite-save-files/topk-analysis/analyze_train_hit.ipynb
% k=1: hit = 22137/(20273*3) =  0.3639816504710699
% k=3: hit = 54262/(40546*3) =  0.4460941482102632
% k=5: hit = 90560/(60819*3) =  0.496336123031728
% k=7: hit = 126858/(81092*3) =  0.5214571104424605
% k=9: hit = 163156/(101365*3) =  0.5365297028888999

% \subsection{Different models: gemma}
% gemma is quite good, ours > gold and weak

% \begin{figure*}
% \includegraphics[width=1\textwidth]{emnlp2023-latex/figures/analysis-data-volume.pdf}
% \caption{}
% \label{fig: analysis-data-volume}
% \end{figure*}

\subsection{Effect of Data Volume in Preference Optimization}
We randomly sample rewrite pairs from those used in the main experiment at various proportions, 20\%$-$80\% with an interval of 20\%, and then use DPO to tune \texttt{QReCC-SFT} on these sets of preference pairs individually. The average performance on QReCC (in-domain) and Doc2dial (out-of-domain) is plotted in Fig. \ref{fig: analysis-data-volume}(a). The performance generally improves with larger data volume. Notably, even with only $20\%$ of pairs, our approach still achieves satisfactory improvement over its SFT version. 

% AdaQR allows us to incorporate additional instances in QReCC training set that lack gold passage labels for preference optimization, while these examples cannot be used for training of \texttt{Gold-Label} approach. To ascertain the benefit of these unlabeled data enabled by our weak supervision, we collect the pair from each example exhibiting the largest reward difference for both \texttt{Gold-Label} and \texttt{Ours}\footnote{Even with the same $\delta$ value, \texttt{Gold-Label} and \texttt{Ours} would obtain different numbers of preference pairs for a given instance, due to the difference in their reward calculation. Here we only collect the pair with the largest reward difference for each instance, instead of all pairs with reward difference greater than $\delta$ as in §\ref{Preference Optimization}, avoiding the variation in the number of resulting preference pairs.}, and  gradually decrease the volume of data used for \texttt{Ours}. As illustrated in Fig. \ref{fig: analysis-data-volume}(b), \texttt{Ours} with $50\%$ of pairs reaches the similar level of performance as \texttt{Gold-Label}, at which point they have a similar number of training pairs. Importantly, when trained with $>50\%$ of pairs, \texttt{Ours} consistently surpasses \texttt{Gold-Label} and peaks with full data, \textbf{highlighting our approach's advantage of exploiting unlabeled data}.
AdaQR allows us to incorporate extra instances in QReCC training set that lack gold passage labels for preference optimization, while these examples cannot be used for training of \texttt{Gold-Label} approach. To verify the benefit of these unlabeled data enabled by our weak supervision, we collect the pair from each example with the largest reward gap for both \texttt{Gold-Label} and \texttt{Ours}\footnote{Even with the same $\delta$ value, \texttt{Gold-Label} and \texttt{Ours} would obtain different numbers of preference pairs for a given instance, due to the difference in their reward calculation. Here we only collect the pair with the largest reward gap for each instance, instead of all pairs with reward gap greater than $\delta$ in §\ref{Preference Optimization}, avoiding the variation in the number of preference pairs.}, and  gradually reduce the size of data used for \texttt{Ours}. In Fig.~\ref{fig: analysis-data-volume}(b), \texttt{Ours} with $40\%$ of pairs reaches the similar level of performance as \texttt{Gold-Label}, at which point \texttt{Ours}  uses 22\% less number of training pairs than \texttt{Gold-Label}. Crucially, when trained on $\geq50\%$ of pairs, \texttt{Ours} consistently exceeds \texttt{Gold-Label} and peaks with full data, \textbf{highlighting AdaQR's advantage of exploiting unlabeled data}.

\section{Related Works}
% \noindent \textbf{Conversational Retrieval with Query Rewriting} 
\noindent \textbf{Conversational Retrieval} 
% Conversational retrieval 
is a precursor task to open-domain conversational question answering. Many existing approaches \cite{Yu_2021, li-etal-2022-grounded,lin2021contextualized} 
% tackles this task by fine-tuning 
fine-tune specialized dense retrievers. However, to leverage the benefits of \textit{off-the-shelf} single-turn retrievers, conversational query rewriting has been applied to transform each conversational question into a standalone query \cite{elgohary-etal-2019-unpack, vakulenko2020question, yu2020fewshot}. Previous approaches typically train query rewriting models with human \cite{elgohary-etal-2019-unpack, anantha-etal-2021-open} or LLM-based \cite{ye-etal-2023-enhancing, jang2024itercqr} rewrite labels. 
% \citet{wu2022conqrr} further trains fine-tuned rewriters using reinforcement learning, with the ranking of passage having maximum similarity to answers as retrievers' preferences. 
% % uses the ranking of passage having maximum similarity to answers as retrievers' preferences to optimize the query rewriting model undergone supervised fine-tuning by reinforcement learning. 
% \citet{mo2024convgqr} trains query rewriting model and query expansion model with an auxiliary loss to minimize the Mean Squared Error (MSE) between embeddings of query and relevant passage. \citet{yoon2024ask} fine-tunes Llama2-7b \cite{touvron2023llama} to replicate the rewrite with the best retrieval feedback, and then aligned it with retrievers' preferences using Direct Preference Optimization (DPO; \citealp{rafailov2024direct}). 
However, acquiring in-domain labels for training on each specific dataset proves costly and the standard fine-tuning alone fails to incorporate retrieval feedback, potentially resulting in sub-optimal performance \cite{wu2022conqrr,yoon2024ask}. Although recent studies \cite{wu2022conqrr, mo2024convgqr, yoon2024ask} suggest integrating signals from retrievers with preference alignment, they require large amounts of in-domain labels. 
% Although these studies suggest integrating signals from retrievers to enhance performance, they necessitates large amount of in-domain rewrite and/or passage labels.
% However, acquiring in-domain labels for training on each specific dataset proves costly and the standard fine-tuning alone fails to incorporate retrieval feedback, potentially resulting in sub-optimal performance \cite{wu2022conqrr,yoon2024ask}. Although recent studies \cite{wu2022conqrr, mo2024convgqr, yoon2024ask} suggest integrating signals from retrievers with preference alignment, they necessitates large amount of in-domain labels. 
% On the contrary, 
To alleviate this data bottleneck, we propose to train effective query rewriting models and assess the adaptation performance with only limited rewrite labels and completely no passage annotation.
\\
\noindent\textbf{Preference Optimization} is a critical research area focused on ensuring that large language models adhere to human values and intents \cite{Bai2022TrainingAH, Ouyang2022TrainingLM, rafailov2024direct}. \citet{Ouyang2022TrainingLM} fine-tunes LLMs with human feedback to align them with user intent, involving collecting human-annotated demonstrations and rankings of model outputs, followed by supervised learning and reinforcement learning. \citet{kim-etal-2023-aligning} uses synthetic feedback instead of human annotations, by reward modeling with synthetic feedback to simulate high-quality demonstrations. Besides improving general abilities of LLMs, some work also focuses on specific aspects. For example, \citet{tian2023finetuning} uses truthfulness measurements as a proxy preference signal to encourage factuality in the model. \citet{yoon2024ask} utilizes the gold passage's ranking as the retrievers’ preferences to optimize the model for rewriting search queries. Our approach is similar to \citet{yoon2024ask} in the target task, query rewriting. However, the preference signals used for aligning our model are synthesized without any retrieval-related labeled data.

\section{Conclusion}
We introduce AdaQR for enhancing query rewriters with minimal to zero in-domain labels,  through the preference optimization towards retrievers' preferences. 
% AdaQR novelly measures 
A novel feature of AdaQR is in measuring
retrievers' preferences with marginal probabilities of answers based on conversations and retrieved passages, 
% allowing us to improve or adapt query rewriters without labeled data. 
enabling improvement or adaptation of query rewriters without labeled data.
Experiments show that AdaQR brings significant improvement to query rewriters' in-domain performance and adapts them well to out-of-domain conversational question answering tasks. AdaQR shows promise in establishing effective query rewriters for arbitrary conversational question answering tasks with minimal effort.

\section*{Limitations}
Although AdaQR demonstrates notable generalization and adaptation capabilities with limited rewrite label requirements, there are still some limitations. 
First, our evaluation in this study was conducted on four datasets. We used QReCC and TopiOCQA as two separate seed datasets, and adapted SFT rewriters to the other three datasets as out-of-domain evaluations. This may not encompass all possible scenarios. 
Secondly, due to computational constraints, we did not engage in detailed analysis of how the quality of rewrite labels and quantity of annotations in the supervised fine-tuning stage might affect the overall performance. Our objective is to use a small amount of labels to attain good retrieval performance.
% as our primary focus was on reducing the required annotations, we did not engage in detailed analysis of how the quality of rewrite labels in the supervised fine-tuning stage may affect the overall performance. 
Further exploration into the impact of label quality and quantity remains an avenue for potential enhancement of rewriting performance and to reduce demands on annotation.
% lessening on annotation demands.
Lastly, we propose AdaQR to derive retrievers' preference using the conditional probability of answers. While we briefly analyzed the impact of different types of passage organization (concatenation and marginalization) in Appendix \ref{sec: appendix-concate}, our primary focus in this paper lies in the thorough analysis of a marginalization approach due to its robustness and effectiveness. Nevertheless, we acknowledge the potential benefit of delving deeper into concatenation-based methods, which might offer valuable insights for the research community in tackling the query rewriting task.

\section*{Ethics Statement}
Query rewriting is instrumental in clarifying users' search intents during information-seeking conversations, improving the retrieval of relevant passages. Our work can greatly enhance the performance of query rewriters. Nevertheless, it is important to recognize that our approach cannot always guarantee perfect rewrites and 
% has a chance of retrieving 
may retrieve 
irrelevant or even nonfactual information. This is partly due to the inherent shortcomings of large language models, which serve as the foundation models in our approach. These models have a propensity to generate hallucinations. Other potential reasons include imperfect retrievers and limited search scopes. 
% These resulting l
Low-quality retrieval result may confuse or even mislead users. Therefore, to ensure the reliability of retrieval results in practical scenarios, it is crucial to implement effective filtering mechanisms, such as rerankers, to identify and exclude passages containing nonfactual information. 

% \newpage
\bibliography{custom}
\bibliographystyle{acl_natbib}

\appendix
\newpage

\section{Prompts}
\label{sec:appendix-prompt}
\noindent \textbf{Supervised Fine-Tuning Label Collection} Table \ref{tab: rewrite-label-prompt} presents the prompt used for rewrite generation on TopiOCQA dataset (§\ref{Supervised Fine-Tuning}). 4278 training instances are derived with Azure OpenAI \texttt{gpt-4 (0314)}. The in-context learning examples are from the QReCC dataset. We do not use in-domain demonstrations for TopiOCQA rewrite label generation since our approach can train effective rewrite models with a limited number of rewrite instances without requiring optimal labels. For QReCC, we use 3850 rewrite labels provided by \citet{ye-etal-2023-enhancing}, who offering annotations in both few-shot learning (FSL) setting and advanced editor setting. 
In the editor setting, ChatGPT refines the rewrites from FSL, functioning as a rewrite editor to provide more competitive results. 
We use the labels generated in the initial FSL setting.

\noindent \textbf{GPT4o Prompting Baselines} Tables \ref{tab: gpt4o-prompt-0shot} and \ref{tab: gpt4o-prompt-1shot} list the prompt for GPT4o prompting used as Doc2Dial and MultiDoc2Dial baselines in Table \ref{tab: main-d2d-md2d}.
\input{prompts}

\section{Experimental Details}
\subsection{Retrieval}
\label{sec:appendix-retrieval}
For sparse retrieval BM25, we use Pyserini \cite{lin2021pyserini} for efficient search and set $k_1=0.82$ and $b=0.68$ in QReCC, and $k_1=0.9$ and $b=0.4$ in TopiOCQA, Doc2Dial and MultiDoc2Dial respectively. $k_1$ controls the non-linear term frequency normalization and $b$ is the scale of the inverse document frequency. For dense retrieval, we use Faiss \cite{johnson2017billionscale} with Exact Search for Inner Product (IndexFlatIP). We employ ANCE \cite{xiong2020approximate} across all dataset except Doc2Dial, with checkpoint trained on MS-MARCO \cite{bajaj2018ms} passage retrieval tasks\footnote{\url{https://huggingface.co/sentence-transformers/msmarco-roberta-base-ance-firstp}}, aligning with previous studies \cite{jang2024itercqr,yoon2024ask}. Our evaluation on dense retrieval of Doc2Dial employs E5-unsupervised \cite{wang2024text} following \citet{liu2024chatqa}\footnote{\url{https://huggingface.co/intfloat/e5-base-unsupervised}}. We set the maximum sequence length to 512 for both ANCE and E5-unsupervised
 We employ the \texttt{pytrec\_eval} toolkit \cite{10.1145/3209978.3210065} for retrieval metric values computation.
 
\subsection{Training}
\label{sec:appendix-training}
We use AdamW \cite{Loshchilov2017DecoupledWD} optimizer with learning rates of 1e-4 and 1e-5 for SFT, DPO stages respectively. The learning rates of both  stages undergo a warmup of $10\%$ of overall training steps, followed by a linear decrease until 0. We set the hyperparameter $\delta$ = 0.1 for organizing preference pairs and $\beta$=0.1 during DPO stage. 

We resort to quantized LoRA (QLoRA) \cite{Hu2021LoRALA, Dettmers2023QLoRAEF} as the parameter-efficient fine-tuning technique for training our models with an NVIDIA A6000 GPU. Specifically, QLoRA is applied to query and value attention matrices inside each decoder block with a fixed rank of 8, a scaling factor of 16, and a dropout probability of 0.05. The model weights are loaded in 4-bit NormalFloat Quantization.

\subsection{Evaluation}

\label{sec:appendix-evaluation}
We train query rewriting models with instances that are not first-turn queries, as these are typically self-contained in application. To ensure fair comparisons with previous baselines during evaluation, we incorporate all first-turn query test instances for both QReCC and TopiOCQA benchmarks. For TopiOCQA, we use the original questions as search inputs. Following previous works \cite{wu2022conqrr, ye-etal-2023-enhancing, anantha-etal-2021-open}, we replace all first user queries in QReCC conversations with their corresponding human rewrites as retrieval queries.
% since some questions are ambiguous in this dataset, requiring additional topical information. 
This step is necessary due to the ambiguity of some questions in this dataset, necessitating additional topical information.
Consequently, the performance of first-turn instances remains consistent across experiments within our setup, i.e., \texttt{SFT}, \texttt{Gold-Label}, \texttt{Pseudo-Label} and \texttt{Ours}. We present results for benchmarks Doc2Dial and MultiDoc2Dial on non-first-turn test instances.
% \section{Results of Gemma and Llama}
% \label{sec:appendix-gemma-llama}
% 
% \section{Analysis of $K$ Values}
% \label{sec:appendix-topk}

\section{Data Statistics}
We list the data statistics of four benchmarks in Table \ref{tab: appendix-data}. As described in Appendix \ref{sec:appendix-evaluation}, we train query rewriting models with instances that are not first-turn queries. During the preference optimization, conversational answers are needed to calculate the marginal probability as reward. The number of evaluation instances for QReCC, TopiOCQA, Doc2Dial and MultiDoc2Dial are $8209$, $2514$, $640$, and $648$ respectively.

\section{Analysis of Topic Shift}
\label{sec:appendix-topic-shift}
\begin{figure}[ht]
\centering
\includegraphics[width=0.43\textwidth]{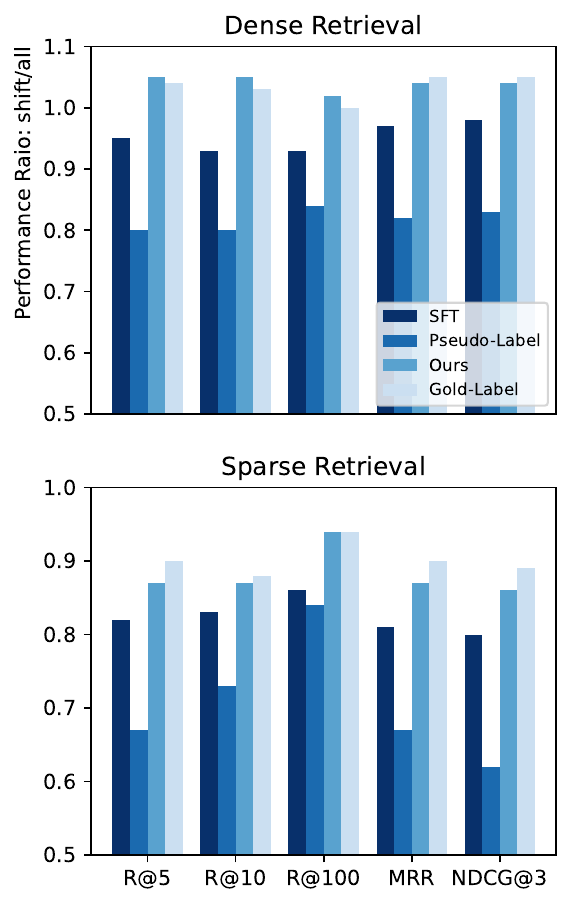}
\caption{Performance ratio of turns with topic-shift and all instances on the test set of TopiOCQA with dense and sparse retrievers. }
\label{fig: analysis-topic-shift}
\end{figure}

Figure \ref{fig: analysis-topic-shift} visualizes the results of topic-shift instances in TopiOCQA, measured by the performance ratio of topic-shift turns over the overall performance. A test example is considered \textit{topic-shift} if the gold passage of the current question differs from the latest history turn. The statistics for initial turns, topic-shift turns and topic-concentrated turns are 205, 672 and 1637 respectively. \textbf{AdaQR demonstrates significant improvement in handling challenging topic-shift instances compared to \texttt{SFT} and \texttt{Pseudo-Label}.} \texttt{Pseudo-Label}'s heavy reliance on word-level overlap between answers and passages not only decreases the overall performance, but also impairs its capability to handle examples involving topic changes. We achieve a performance ratio comparable to the \texttt{Gold-Label} counterpart and even surpasses it on certain metrics with a dense retriever.

\section{Analysis of Passage Concatenation and Marginalization}
\label{sec: appendix-concate}
\input{appendix-concate}

To fully assess the effect of using the conditional probability of answers as retrieval preference, we introduce and compare two types of passage organization designs: (1) \texttt{M5} follows the description in §\ref{Reward Collection} that computes the probability by marginalizing top-$K$ passages. (2) \texttt{C5} is proposed as a variation to \texttt{M5}, which computes the probability of answers conditioned on concatenated top-$K$ passages as a single input. The retrieval performance with top-$5$ and top-$9$ passages are reported in Table \ref{tab: appendix-concate}. In general, \texttt{M5} surpasses \texttt{C5} on the average performance and metrics with relatively small @$k$ values. On QReCC, we observed that the improvement percentage increases when the inclusion of more passages during the reward calculation, i.e., $9$ vs. $5$. This suggest that \texttt{M5} may exhibit greater resilience to noise in information. Nevertheless, \texttt{C5} demonstrates its own advantage by achieving relatively strong performance with fewer input tokens during reward calculation.
% \newpage

\input{algorithm}
\input{appendix-data}

\input{appendix-gemma-llama}
\input{analysis-top-k}

\input{appendix-data-volume}

\end{document}

%% file: main-qrecc-topiocqa.tex
\begin{table*}[ht]
\centering
\scalebox{0.65}{
\begin{tabular}{clccccccccccccc}
\toprule
\multicolumn{1}{l}{} & \multicolumn{8}{c|}{\textbf{QReCC (8209)}} & \multicolumn{6}{c}{\textbf{TopiOCQA (2514)}} \\
\multicolumn{1}{l}{Type} & Method & MRR & MAP & NDCG & R@1 & R@5 & R@10 & \multicolumn{1}{c|}{R@50} & MRR & NDCG & R@1 & R@5 & R@10 & R@100 \\ \midrule \midrule
\multicolumn{1}{c|}{} & T5QR* & 33.4 & - & 30.2 & - & - & 53.8 & \multicolumn{1}{c|}{-} & 11.3 & 9.8 & - & - & 22.1 & 44.7 \\
\multicolumn{1}{c|}{} & CONQRR* & 38.3 & - & - & - & - & 60.1 & \multicolumn{1}{c|}{-} & - & - & - & - & - & - \\
\multicolumn{1}{c|}{} & \cellcolor[HTML]{ECF4FF}ConvGQR* & \cellcolor[HTML]{ECF4FF}44.1 & \cellcolor[HTML]{ECF4FF}- & \cellcolor[HTML]{ECF4FF}41.0 & \cellcolor[HTML]{ECF4FF}- & \cellcolor[HTML]{ECF4FF}- & \cellcolor[HTML]{ECF4FF}64.4 & \multicolumn{1}{c|}{\cellcolor[HTML]{ECF4FF}-} & \cellcolor[HTML]{ECF4FF}12.4 & \cellcolor[HTML]{ECF4FF}10.7 & \cellcolor[HTML]{ECF4FF}- & \cellcolor[HTML]{ECF4FF}- & \cellcolor[HTML]{ECF4FF}23.8 & \cellcolor[HTML]{ECF4FF}45.6 \\
\multicolumn{1}{c|}{} & \cellcolor[HTML]{ECF4FF}IterCQR* & \cellcolor[HTML]{ECF4FF}46.7 & \cellcolor[HTML]{ECF4FF}- & \cellcolor[HTML]{ECF4FF}44.1 & \cellcolor[HTML]{ECF4FF}- & \cellcolor[HTML]{ECF4FF}- & \cellcolor[HTML]{ECF4FF}64.4 & \multicolumn{1}{c|}{\cellcolor[HTML]{ECF4FF}-} & \cellcolor[HTML]{ECF4FF}16.5 & \cellcolor[HTML]{ECF4FF}14.9 & \cellcolor[HTML]{ECF4FF}- & \cellcolor[HTML]{ECF4FF}- & \cellcolor[HTML]{ECF4FF}29.3 & \cellcolor[HTML]{ECF4FF}54.1 \\
\multicolumn{1}{c|}{} & LLM IQR* & 49.4 & 47.9 & 56.9 & 36.4 & 58.9 & 67.0 & \multicolumn{1}{c|}{83.1} & - & - & - & - & - & - \\
\multicolumn{1}{c|}{} & GPT4 Prompting* & - & - & - & - & - & - & \multicolumn{1}{c|}{-} & 18.5 & - & - & - & 35.1 & 62.9 \\
\multicolumn{1}{c|}{} & \cellcolor[HTML]{ECF4FF}Llama2 Distill* & \cellcolor[HTML]{ECF4FF}- & \cellcolor[HTML]{ECF4FF}- & \cellcolor[HTML]{ECF4FF}- & \cellcolor[HTML]{ECF4FF}- & \cellcolor[HTML]{ECF4FF}- & \cellcolor[HTML]{ECF4FF}- & \multicolumn{1}{c|}{\cellcolor[HTML]{ECF4FF}-} & \cellcolor[HTML]{ECF4FF}19.0 & \cellcolor[HTML]{ECF4FF}- & \cellcolor[HTML]{ECF4FF}- & \cellcolor[HTML]{ECF4FF}- & \cellcolor[HTML]{ECF4FF}35.5 & \cellcolor[HTML]{ECF4FF}64.6 \\
\multicolumn{1}{c|}{} & \cellcolor[HTML]{ECF4FF}RETPO* & \cellcolor[HTML]{ECF4FF}50.0 & \cellcolor[HTML]{ECF4FF}- & \cellcolor[HTML]{ECF4FF}47.3 & \cellcolor[HTML]{ECF4FF}- & \cellcolor[HTML]{ECF4FF}- & \cellcolor[HTML]{ECF4FF}69.5 & \multicolumn{1}{c|}{\cellcolor[HTML]{ECF4FF}-} & \cellcolor[HTML]{ECF4FF}28.3 & \cellcolor[HTML]{ECF4FF}26.5 & \cellcolor[HTML]{ECF4FF}- & \cellcolor[HTML]{ECF4FF}- & \cellcolor[HTML]{ECF4FF}48.3 & \cellcolor[HTML]{ECF4FF}73.1 \\ \cline{2-15} 
% \multicolumn{1}{c|}{} & \multicolumn{14}{c}{Seed Dataset: QReCC} \\ \cline{2-15} 
\multicolumn{1}{c|}{} & QReCC-SFT & 45.9 & 44.4 & 43.2 & 32.4 & 56.0 & 64.9 & \multicolumn{1}{c|}{83.7} & 16.3 & 14.6 & 10.2 & 22.4 & 29.3 & 52.1 \\
% \multicolumn{1}{c|}{} & Pseudo-Label & 50.5 & 49.0 & 48.1 & 37.0 & 60.8 & 69.4 & \multicolumn{1}{c|}{86.1} & 13.1 & 11.6 & 8.1 & 17.4 & 23.4 & 47.2 \\
\multicolumn{1}{c|}{} & \cellcolor[HTML]{ECF4FF}\ + Gold-Label& \cellcolor[HTML]{ECF4FF}51.9 & \cellcolor[HTML]{ECF4FF}50.3 & \cellcolor[HTML]{ECF4FF}49.4 & \cellcolor[HTML]{ECF4FF}38.8 & \cellcolor[HTML]{ECF4FF}\textbf{61.6} & \cellcolor[HTML]{ECF4FF}\textbf{69.8} & \multicolumn{1}{c|}{\cellcolor[HTML]{ECF4FF}\textbf{86.8}} & \cellcolor[HTML]{ECF4FF}\textbf{20.6} & \cellcolor[HTML]{ECF4FF}18.5 & \cellcolor[HTML]{ECF4FF}12.8 & \cellcolor[HTML]{ECF4FF}\textbf{28.7} & \cellcolor[HTML]{ECF4FF}\textbf{37.2} & \cellcolor[HTML]{ECF4FF}\textbf{65.1} \\
\multicolumn{1}{c|}{} & \textbf{\ + Ours} & \textbf{52.3} & \textbf{50.8} & \textbf{49.9} & \textbf{39.8} & 61.3 & 69.1 & \multicolumn{1}{c|}{85.0} & 20.5 & \textbf{18.9} & \textbf{13.4} & 27.8 & 34.8 & 61.3 \\ \cline{2-15}  
% \multicolumn{1}{c|}{} & \multicolumn{14}{c}{Seed Dataset: TopiOCQA} \\ \cline{2-15} 
\multicolumn{1}{c|}{} & TopiOCQA-SFT & 40.8 & 39.3 & 37.6 & 26.6 & 51.6 & 62.4 & \multicolumn{1}{c|}{83.4} & 17.7 & 15.5 & 9.9 & 25.7 & 34.4 & 62.0 \\
% \multicolumn{1}{c|}{} & Pseudo-Label & 47.8 & 46.2 & 45.0 & 33.4 & 59.1 & 68.7 & \multicolumn{1}{c|}{\textbf{87.2}} & 16.8 & 14.8 & 10.6 & 23.0 & 30.0 & 56.3 \\
\multicolumn{1}{c|}{} & \cellcolor[HTML]{ECF4FF}\ + Gold-Label& \cellcolor[HTML]{ECF4FF}48.5 & \cellcolor[HTML]{ECF4FF}47.0 & \cellcolor[HTML]{ECF4FF}45.9 & \cellcolor[HTML]{ECF4FF}34.5 & \cellcolor[HTML]{ECF4FF}59.0 & \cellcolor[HTML]{ECF4FF}68.6 & \multicolumn{1}{c|}{\cellcolor[HTML]{ECF4FF}\textbf{87.2}} & \cellcolor[HTML]{ECF4FF}\textbf{20.5} & \cellcolor[HTML]{ECF4FF}\textbf{18.1} & \cellcolor[HTML]{ECF4FF}\textbf{12.3} & \cellcolor[HTML]{ECF4FF}\textbf{29.0} & \cellcolor[HTML]{ECF4FF}\textbf{38.2} & \cellcolor[HTML]{ECF4FF}\textbf{68.0} \\
\multicolumn{1}{c|}{\multirow{-16}{*}{\begin{tabular}[c]{@{}c@{}}\rotatebox[origin=c]{90}{Sparse Retrieval}\end{tabular}}} & \textbf{\ + Ours} & \textbf{50.6} & \textbf{49.0} & \textbf{48.0} & \textbf{37.0} & \textbf{60.7} & \textbf{69.6} & \multicolumn{1}{c|}{86.7} & 20.3 & 18.0 & \textbf{12.3} & 28.2 & 37.1 & 66.2 \\ \midrule \midrule
\multicolumn{1}{c|}{} & T5QR* & 34.5 & - & 31.8 & - & - & 53.1 & \multicolumn{1}{c|}{-} & 23.0 & 22.2 & - & - & 37.6 & 54.4 \\
\multicolumn{1}{c|}{} & CONQRR* & 41.8 & - & - & - & - & 65.1 & \multicolumn{1}{c|}{-} & - & - & - & - & - & - \\
\multicolumn{1}{c|}{} & \cellcolor[HTML]{ECF4FF}ConvGQR* & \cellcolor[HTML]{ECF4FF}42.0 & \cellcolor[HTML]{ECF4FF}- & \cellcolor[HTML]{ECF4FF}39.1 & \cellcolor[HTML]{ECF4FF}- & \cellcolor[HTML]{ECF4FF}- & \cellcolor[HTML]{ECF4FF}63.5 & \multicolumn{1}{c|}{\cellcolor[HTML]{ECF4FF}-} & \cellcolor[HTML]{ECF4FF}25.6 & \cellcolor[HTML]{ECF4FF}24.3 & \cellcolor[HTML]{ECF4FF}- & \cellcolor[HTML]{ECF4FF}- & \cellcolor[HTML]{ECF4FF}41.8 & \cellcolor[HTML]{ECF4FF}58.8 \\
\multicolumn{1}{c|}{} & \cellcolor[HTML]{ECF4FF}IterCQR* & \cellcolor[HTML]{ECF4FF}42.9 & \cellcolor[HTML]{ECF4FF}- & \cellcolor[HTML]{ECF4FF}40.2 & \cellcolor[HTML]{ECF4FF}- & \cellcolor[HTML]{ECF4FF}- & \cellcolor[HTML]{ECF4FF}65.5 & \multicolumn{1}{c|}{\cellcolor[HTML]{ECF4FF}-} & \cellcolor[HTML]{ECF4FF}26.3 & \cellcolor[HTML]{ECF4FF}25.1 & \cellcolor[HTML]{ECF4FF}- & \cellcolor[HTML]{ECF4FF}- & \cellcolor[HTML]{ECF4FF}42.6 & \cellcolor[HTML]{ECF4FF}62.0 \\
\multicolumn{1}{c|}{} & InstructLLM* & 43.5 & - & 40.5 & - & - & 66.7 & \multicolumn{1}{c|}{-} & 25.3 & 23.7 & - & - & 45.1 & 69.0 \\
\multicolumn{1}{c|}{} & \cellcolor[HTML]{ECF4FF}RETPO* & \cellcolor[HTML]{ECF4FF}44.0 & \cellcolor[HTML]{ECF4FF}- & \cellcolor[HTML]{ECF4FF}41.1 & \cellcolor[HTML]{ECF4FF}- & \cellcolor[HTML]{ECF4FF}- & \cellcolor[HTML]{ECF4FF}66.7 & \multicolumn{1}{c|}{\cellcolor[HTML]{ECF4FF}-} & \cellcolor[HTML]{ECF4FF}30.0 & \cellcolor[HTML]{ECF4FF}28.9 & \cellcolor[HTML]{ECF4FF}- & \cellcolor[HTML]{ECF4FF}- & \cellcolor[HTML]{ECF4FF}49.6 & \cellcolor[HTML]{ECF4FF}68.7 \\ \cline{2-15} 
% \multicolumn{1}{c|}{} & \multicolumn{14}{c}{Seed Dataset: QReCC} \\ \cline{2-15} 
\multicolumn{1}{c|}{} & QReCC-SFT & 41.2 & 39.4 & 38.5 & 27.1 & 52.7 & 61.9 & \multicolumn{1}{c|}{76.4} & 25.6 & 24.2 & 16.4 & 37.0 & 43.8 & 63.5 \\
% \multicolumn{1}{c|}{} & Pseudo-Label & \textbf{45.5} & \textbf{43.6} & 42.8 & 30.4 & 58.1 & \textbf{67.9} & \multicolumn{1}{c|}{\textbf{82.1}} & 21.9 & 20.2 & 13.6 & 31.3 & 38.7 & 61 \\
\multicolumn{1}{c|}{} & \cellcolor[HTML]{ECF4FF}\ + Gold-Label & \cellcolor[HTML]{ECF4FF}\textbf{45.5} & \cellcolor[HTML]{ECF4FF}\textbf{43.5} & \cellcolor[HTML]{ECF4FF}\textbf{42.8} & \cellcolor[HTML]{ECF4FF}\textbf{30.4} & \cellcolor[HTML]{ECF4FF}\textbf{58.3} & \cellcolor[HTML]{ECF4FF}\textbf{67.7} & \multicolumn{1}{c|}{\cellcolor[HTML]{ECF4FF}\textbf{81.5}} & \cellcolor[HTML]{ECF4FF}\textbf{36.4} & \cellcolor[HTML]{ECF4FF}\textbf{35.2} & \cellcolor[HTML]{ECF4FF}24.3 & \cellcolor[HTML]{ECF4FF}\textbf{51.2} & \cellcolor[HTML]{ECF4FF}\textbf{59.8} & \cellcolor[HTML]{ECF4FF}\textbf{79.6} \\
\multicolumn{1}{c|}{} & \textbf{\ + Ours} & 45.3 & \textbf{43.5} & 42.7 & \textbf{30.4} & 58.1 & 67.2 & \multicolumn{1}{c|}{81.4} & 36.0 & 34.6 & \textbf{24.5} & 50.5 & 58.2 & 78.7 \\ \cline{2-15}
% \multicolumn{1}{c|}{} & \multicolumn{14}{c}{Seed Dataset: TopiOCQA} \\ \cline{2-15} 
\multicolumn{1}{c|}{} & TopiOCQA-SFT & 39.8 & 37.8 & 36.9 & 25.8 & 50.9 & 60.4 & \multicolumn{1}{c|}{75.7} & 33.4 & 31.9 & 22.8 & 46.7 & 54.7 & 73.8 \\
% \multicolumn{1}{c|}{} & Pseudo-Label & 43.3 & 41.4 & 40.5 & 27.8 & \textbf{56.0} & \textbf{66.5} & \multicolumn{1}{c|}{\textbf{81.6}} & 29.0 & 27.4 & 19.0 & 40.2 & 49.5 & 71.3 \\
\multicolumn{1}{c|}{} & \cellcolor[HTML]{ECF4FF}\ + Gold-Label & \cellcolor[HTML]{ECF4FF}43.2 & \cellcolor[HTML]{ECF4FF}41.2 & \cellcolor[HTML]{ECF4FF}40.4 & \cellcolor[HTML]{ECF4FF}28.0 & \cellcolor[HTML]{ECF4FF}\textbf{55.8} & \cellcolor[HTML]{ECF4FF}\textbf{65.8} & \multicolumn{1}{c|}{\cellcolor[HTML]{ECF4FF}\textbf{80.9}} & \cellcolor[HTML]{ECF4FF}37.5 & \cellcolor[HTML]{ECF4FF}36.1 & \cellcolor[HTML]{ECF4FF}25.8 & \cellcolor[HTML]{ECF4FF}51.8 & \cellcolor[HTML]{ECF4FF}60.7 & \cellcolor[HTML]{ECF4FF}79.8 \\
\multicolumn{1}{c|}{\multirow{-12}{*}{\begin{tabular}[c]{@{}c@{}}\rotatebox[origin=c]{90}{Dense Retrieval}\end{tabular}}} & \textbf{\ + Ours} & \textbf{43.4} & \textbf{41.5} & \textbf{40.8} & \textbf{28.3} & \textbf{55.8} & 65.6 & \multicolumn{1}{c|}{80.4} & \textbf{38.1} & \textbf{36.6} & \textbf{26.3} & \textbf{53.0} & \textbf{61.3} & \textbf{79.9} \\ \bottomrule
\end{tabular}}
\caption{Evaluation results of sparse and dense retrieval on QReCC and TopiOCQA. Two SFT models (\texttt{QReCC-SFT} and \texttt{TopiOCQA-SFT}) are evaluated to demonstrate the in-domain and out-of-domain performance. 
We include baselines 
% with query reformulation models 
following \citet{yoon2024ask} and \citet{ye-etal-2023-enhancing}, denoted with *. 
% InstructLLM fine-tuned the retriever's query encoder. 
Methods requiring in-domain passage labels are marked with \colorbox[HTML]{ECF4FF}{background color} in this and subsequent tables. 
% For fair comparison to the baselines, we include all all instances of first-turn queries with original questions as search inputs. Hence the performance of first-turn instances are the same for \texttt{SFT}, \texttt{Gold-Label} and \texttt{Ours}. 
% Results with Gemma-7B and Llama2-7B are in Appendix Table \ref{tab:gemma-llama}.
See experimental details in Appendix \ref{sec:appendix-evaluation}.}
\label{tab: main-qrecc-topiocqa}
\vspace{-0.5\baselineskip}
\end{table*}

%% file: main-d2d-md2d.tex
\begin{table}[ht]
% \centering
\scalebox{0.62}{
\begin{tabular}{clcccccc}
\toprule
\multicolumn{8}{c}{\textbf{Doc2dial (640)}} \\
Type & \multicolumn{1}{c}{Method} & MRR & NDCG & R@1 & R@5 & R@10 & R@20 \\ \midrule \midrule
\multicolumn{1}{c|}{} & GPT4o 0-shot & 51.8 & 51.1 & 37.8 & 69.8 & 77.0 & 83.6 \\
\multicolumn{1}{c|}{} & GPT4o 1-shot & 53.8 & 53.2 & 40.2 & 70.0 & 77.8 & 86.7 \\ \cline{2-8} 
\multicolumn{1}{c|}{} & QReCC-SFT & 56.0 & 55.7 & 42.3 & 72.0 & 81.7 & 89.5 \\
\multicolumn{1}{c|}{} & \cellcolor[HTML]{ECF4FF}+ Gold-Label & \cellcolor[HTML]{ECF4FF}\textbf{60.6} & \cellcolor[HTML]{ECF4FF}\textbf{60.6} & \cellcolor[HTML]{ECF4FF}\textbf{46.1} & \cellcolor[HTML]{ECF4FF}\textbf{78.0} & \cellcolor[HTML]{ECF4FF}\textbf{85.2} & \cellcolor[HTML]{ECF4FF}\textbf{91.4} \\
\multicolumn{1}{c|}{} & \textbf{+ Ours} & 59.9 & 59.7 & \textbf{46.1} & 77.3 & 84.5 & 90.0 \\ \cline{2-8} 
\multicolumn{1}{c|}{} & TopiOCQA-SFT & 56.4 & 55.9 & 41.9 & 74.5 & 82.3 & 88.8 \\
\multicolumn{1}{c|}{} & \cellcolor[HTML]{ECF4FF}+ Gold-Label & \cellcolor[HTML]{ECF4FF}\textbf{62.1} & \cellcolor[HTML]{ECF4FF}\textbf{62.1} & \cellcolor[HTML]{ECF4FF}47.5 & \cellcolor[HTML]{ECF4FF}\textbf{81.1} & \cellcolor[HTML]{ECF4FF}\textbf{87.2} & \cellcolor[HTML]{ECF4FF}\textbf{92.8} \\
\multicolumn{1}{c|}{\multirow{-8}{*}{\rotatebox[origin=c]{90}{Sparse}}} & \textbf{+ Ours} & 61.8 & 61.8 & \textbf{47.7} & 80.5 & 86.3 & 91.4 \\ \midrule
\multicolumn{1}{c|}{} & GPT4o 0-shot & 44.9 & 43.4 & 32.5 & 59.1 & 68.3 & 77.2 \\
\multicolumn{1}{c|}{} & GPT4o 1-shot & 45.6 & 43.9 & 33.0 & 60.3 & 68.9 & 78.9 \\ \cline{2-8} 
\multicolumn{1}{c|}{} & QReCC-SFT & 47.3 & 46.8 & 32.0 & 64.2 & 75.8 & 84.4 \\
\multicolumn{1}{c|}{} & \cellcolor[HTML]{ECF4FF}+ Gold-Label & \cellcolor[HTML]{ECF4FF}\textbf{54.4} & \cellcolor[HTML]{ECF4FF}\textbf{53.8} & \cellcolor[HTML]{ECF4FF}\textbf{38.6} & \cellcolor[HTML]{ECF4FF}\textbf{73.0} & \cellcolor[HTML]{ECF4FF}\textbf{84.8} & \cellcolor[HTML]{ECF4FF}\textbf{91.9} \\
\multicolumn{1}{c|}{} & \textbf{+ Ours} & 53.6 & 52.9 & 38.4 & 71.3 & 81.9 & 91.1 \\ \cline{2-8} 
\multicolumn{1}{c|}{} & TopiOCQA-SFT & 46.5 & 45.1 & 32.3 & 62.7 & 73.8 & 81.4 \\
\multicolumn{1}{c|}{} & \cellcolor[HTML]{ECF4FF}+ Gold-Label & \cellcolor[HTML]{ECF4FF}\textbf{53.9} & \cellcolor[HTML]{ECF4FF}\textbf{53.6} & \cellcolor[HTML]{ECF4FF}\textbf{38.4} & \cellcolor[HTML]{ECF4FF}\textbf{71.6} & \cellcolor[HTML]{ECF4FF}\textbf{82.0} & \cellcolor[HTML]{ECF4FF}\textbf{88.9} \\
\multicolumn{1}{c|}{\multirow{-8}{*}{\rotatebox[origin=c]{90}{Dense}}} & \textbf{+ Ours} & 51.3 & 50.4 & 35.6 & 69.2 & 79.7 & 87.8 \\ \midrule \midrule
\multicolumn{8}{c}{\textbf{MultiDoc2dial (648)}} \\ \midrule
\multicolumn{1}{c|}{} & GPT4o 0-shot & 47.8 & 46.7 & 34.7 & 63.0 & 72.5 & 80.6 \\
\multicolumn{1}{c|}{} & GPT4o 1-shot & 48.7 & 47.9 & 35.0 & 65.6 & 74.2 & 83.3 \\ \cline{2-8} 
\multicolumn{1}{c|}{} & QReCC-SFT & 51.4 & 50.9 & 38.6 & 65.7 & 75.3 & 82.7 \\
\multicolumn{1}{c|}{} & \cellcolor[HTML]{ECF4FF}+ Gold-Label & \cellcolor[HTML]{ECF4FF}\textbf{55.7} & \cellcolor[HTML]{ECF4FF}55.4 & \cellcolor[HTML]{ECF4FF}42.0 & \cellcolor[HTML]{ECF4FF}\textbf{71.9} & \cellcolor[HTML]{ECF4FF}\textbf{81.3} & \cellcolor[HTML]{ECF4FF}\textbf{88.0} \\
\multicolumn{1}{c|}{} & \textbf{+ Ours} & 55.6 & \textbf{55.6} & \textbf{42.1} & 71.5 & 80.3 & 87.0 \\ \cline{2-8} 
\multicolumn{1}{c|}{} & TopiOCQA-SFT & 52.1 & 50.9 & 38.6 & 69.0 & 77.0 & 84.7 \\
\multicolumn{1}{c|}{} & \cellcolor[HTML]{ECF4FF}+ Gold-Label & \cellcolor[HTML]{ECF4FF}55.3 & \cellcolor[HTML]{ECF4FF}54.1 & \cellcolor[HTML]{ECF4FF}42.4 & \cellcolor[HTML]{ECF4FF}70.5 & \cellcolor[HTML]{ECF4FF}80.3 & \cellcolor[HTML]{ECF4FF}\textbf{87.7} \\
\multicolumn{1}{c|}{\multirow{-8}{*}{\rotatebox[origin=c]{90}{Sparse}}} & \textbf{+ Ours} & \textbf{56.6} & \textbf{55.9} & \textbf{43.5} & \textbf{73.0} & \textbf{81.3} & 87.2 \\ \midrule
\multicolumn{1}{c|}{} & GPT4o 0-shot & 39.3 & 37.1 & 27.6 & 54.5 & 64.0 & 71.9 \\
\multicolumn{1}{c|}{} & GPT4o 1-shot & 39.8 & 37.8 & 27.5 & 54.0 & 65.9 & 73.6 \\ \cline{2-8} 
\multicolumn{1}{c|}{} & QReCC-SFT & 41.6 & 40.7 & 27.6 & 57.3 & 67.9 & 75.9 \\
\multicolumn{1}{c|}{} & \cellcolor[HTML]{ECF4FF}+ Gold-Label & \cellcolor[HTML]{ECF4FF}\textbf{45.7} & \cellcolor[HTML]{ECF4FF}\textbf{44.8} & \cellcolor[HTML]{ECF4FF}\textbf{31.5} & \cellcolor[HTML]{ECF4FF}\textbf{62.5} & \cellcolor[HTML]{ECF4FF}\textbf{72.1} & \cellcolor[HTML]{ECF4FF}\textbf{83.5} \\
\multicolumn{1}{c|}{} & \textbf{+ Ours} & 44.3 & 42.9 & 30.3 & 60.5 & 71.3 & 81.3 \\ \cline{2-8} 
\multicolumn{1}{c|}{} & TopiOCQA-SFT & 39.9 & 37.8 & 27.8 & 53.1 & 64.0 & 74.4 \\
\multicolumn{1}{c|}{} & \cellcolor[HTML]{ECF4FF}+ Gold-Label & \cellcolor[HTML]{ECF4FF}\textbf{45.1} & \cellcolor[HTML]{ECF4FF}\textbf{43.3} & \cellcolor[HTML]{ECF4FF}\textbf{31.9} & \cellcolor[HTML]{ECF4FF}\textbf{58.6} & \cellcolor[HTML]{ECF4FF}\textbf{72.8} & \cellcolor[HTML]{ECF4FF}\textbf{82.4} \\
\multicolumn{1}{c|}{\multirow{-8}{*}{\rotatebox[origin=c]{90}{Dense}}} & \textbf{+ Ours} & 43.8 & 42.2 & 30.4 & 57.6 & 70.2 & 80.9 \\ \bottomrule
\end{tabular}}
\caption{Evaluation results of sparse and dense retrieval on Doc2Dial and MultiDoc2Dial. Both two SFT models (\texttt{QReCC-SFT} and \texttt{TopiOCQA-SFT}) are out-of-domain evaluation. We include zero-shot and one-shot (an example from QReCC) learning performance of GPT4o as comparison baselines. See prompt in Appendix \ref{sec:appendix-prompt}. 
% Methods requiring in-domain labels are marked with \colorbox[HTML]{ECF4FF}{background color}.
}
\label{tab: main-d2d-md2d}
\vspace{-1\baselineskip}
\end{table}

%% file: analysis-weak.tex
% \usepackage{multirow}
\begin{table*}[ht]
\centering
\scalebox{0.64}{
\begin{tabular}{c|lcccc|cccc|cccc|cccc}
\toprule
\multicolumn{1}{l|}{} & \multicolumn{5}{c|}{\textbf{QReCC (8209)}} & \multicolumn{4}{c|}{\textbf{TopiOCQA (2514)}} & \multicolumn{4}{c|}{\textbf{Doc2Dial (640)}} & \multicolumn{4}{l}{\textbf{MultiDoc2dial (648)}} \\
Type & Method & MRR & R@5 & R@50 & AVG & MRR & R@5 & R@100 & AVG & MRR & R@5 & R@20 & AVG & MRR & R@5 & R@20 & AVG \\ \midrule \midrule 
\multirow{6}{*}{\rotatebox[origin=c]{90}{Sparse}} & QReCC-SFT & 45.9 & 56.0 & \multicolumn{1}{c|}{83.7} & 61.8 & 16.3 & 22.4 & \multicolumn{1}{c|}{52.1} & 30.3 & 56.0 & 72.0 & \multicolumn{1}{c|}{89.5} & 72.5 & 51.4 & 65.7 & \multicolumn{1}{c|}{82.7} & 66.6 \\
 & + Pseudo-Label & 50.5 & 60.8 & \multicolumn{1}{c|}{\textbf{86.1}} & 65.8 & 13.1 & 17.4 & \multicolumn{1}{c|}{47.2} & 25.9 & 58.0 & 75.8 & \multicolumn{1}{c|}{\textbf{90.0}} & 74.6 & 52.3 & 69.6 & \multicolumn{1}{c|}{85.0} & 69.0 \\
 & + \textbf{Ours} & \textbf{52.3} & \textbf{61.3} & \multicolumn{1}{c|}{85.0} & \textbf{66.2} & \textbf{20.5} & \textbf{27.8} & \multicolumn{1}{c|}{\textbf{61.3}} & \textbf{36.5} & \textbf{59.9} & \textbf{77.3} & \multicolumn{1}{c|}{\textbf{90.0}} & \textbf{75.7} & \textbf{55.6} & \textbf{71.5} & \multicolumn{1}{c|}{\textbf{87.0}} & \textbf{71.4} \\ \cline{2-18} 
 & TopiOCQA-SFT & 40.8 & 51.6 & \multicolumn{1}{c|}{83.4} & 58.6 & 17.7 & 25.7 & \multicolumn{1}{c|}{62.0} & 35.1 & 56.4 & 74.5 & \multicolumn{1}{c|}{88.8} & 73.2 & 52.1 & 69.0 & \multicolumn{1}{c|}{84.7} & 68.6 \\
 & + Pseudo-Label & 47.8 & 59.1 & \multicolumn{1}{c|}{\textbf{87.2}} & 64.7 & 16.8 & 23.0 & \multicolumn{1}{c|}{56.3} & 32.0 & 61.1 & 78.3 & \multicolumn{1}{c|}{90.6} & 76.7 & 54.6 & 71.6 & \multicolumn{1}{c|}{86.0} & 70.7 \\
 & + \textbf{Ours} & \textbf{50.6} & \textbf{60.7} & \multicolumn{1}{c|}{86.7} & \textbf{66.0} & \textbf{20.3} & \textbf{28.2} & \multicolumn{1}{c|}{\textbf{66.2}} & \textbf{38.2} & \textbf{61.8} & \textbf{80.5} & \multicolumn{1}{c|}{\textbf{91.4}} & \textbf{77.9} & \textbf{56.6} & \textbf{73.0} & \multicolumn{1}{c|}{\textbf{87.2}} & \textbf{72.3} \\ \midrule \midrule
\multirow{6}{*}{\rotatebox[origin=c]{90}{Dense}} & QReCC-SFT & 41.2 & 52.7 & \multicolumn{1}{c|}{76.4} & 56.8 & 25.6 & 37.0 & \multicolumn{1}{c|}{63.5} & 42.0 & 47.3 & 64.2 & \multicolumn{1}{c|}{84.4} & 65.3 & 41.6 & 57.3 & \multicolumn{1}{c|}{75.9} & 58.2 \\
 & + Pseudo-Label & \textbf{45.5} & \textbf{58.1} & \multicolumn{1}{c|}{\textbf{82.1}} & \textbf{61.9} & 21.9 & 31.3 & \multicolumn{1}{c|}{61.0} & 38.0 & 52.3 & 70.8 & \multicolumn{1}{c|}{89.5} & 70.9 & 43.1 & 59.1 & \multicolumn{1}{c|}{80.7} & 61.0 \\
 & + \textbf{Ours} & 45.3 & \textbf{58.1} & \multicolumn{1}{c|}{81.4} & 61.6 & \textbf{36.0} & \textbf{50.5} & \multicolumn{1}{c|}{\textbf{78.7}} & \textbf{55.1} & \textbf{53.6} & \textbf{71.3} & \multicolumn{1}{c|}{\textbf{91.1}} & \textbf{72.0} & \textbf{44.3} & \textbf{60.5} & \multicolumn{1}{c|}{\textbf{81.3}} & \textbf{62.1} \\ \cline{2-18} 
 & TopiOCQA-SFT & 39.8 & 50.9 & \multicolumn{1}{c|}{75.7} & 55.5 & 33.4 & 46.7 & \multicolumn{1}{c|}{73.8} & 51.3 & 46.5 & 62.7 & \multicolumn{1}{c|}{81.4} & 63.5 & 39.9 & 53.1 & \multicolumn{1}{c|}{74.4} & 55.8 \\
 & + Pseudo-Label & 43.3 & \textbf{56.0} & \multicolumn{1}{c|}{\textbf{81.6}} & \textbf{60.3} & 29.0 & 40.2 & \multicolumn{1}{c|}{71.3} & 46.8 & \textbf{51.3} & 68.0 & \multicolumn{1}{c|}{\textbf{88.0}} & 69.1 & 43.6 & \textbf{57.6} & \multicolumn{1}{c|}{78.9} & 60.0 \\
 & + \textbf{Ours} & \textbf{43.4} & 55.8 & \multicolumn{1}{c|}{80.4} & 59.9 & \textbf{38.1} & \textbf{53.0} & \multicolumn{1}{c|}{\textbf{79.9}} & \textbf{57.0} & \textbf{51.3} & \textbf{69.2} & \multicolumn{1}{c|}{87.8} & \textbf{69.5} & \textbf{43.8} & \textbf{57.6} & \multicolumn{1}{c|}{\textbf{80.9}} & \textbf{60.7} \\ \bottomrule
\end{tabular}
}
\caption{Comparison between two weakly supervised approaches: \texttt{Pseudo-Label} and \texttt{Ours}. 
% Evaluation results of sparse and dense retrieval on QReCC, TopiOCQA, Doc2Dial and MultiDoc2Dial are listed with two SFT versions, i.e., \texttt{QReCC-SFT} and \texttt{TopiOCQA-SFT}. 
Evaluation results of sparse and dense retrieval with two SFT versions, i.e., \texttt{QReCC-SFT} and \texttt{TopiOCQA-SFT}, are listed.
}
\label{tab:pseudo-vs-ours}
\vspace{-0.5\baselineskip}
\end{table*}

%% file: prompts.tex
% \citet{ye-etal-2023-enhancing} used prompt in Table \ref{tab: rewrite-label-prompt} to generate rewrite labels for QReCC under few-shot learning setting with ChatGPT (\texttt{gpt-3.5-turbo}). We follow the same prompt to derive rewrite labels for TopiOCQA. 
% Note that the four in-context learning examples are from the QReCC dataset, meaning we do not use in-domain demonstrations for TopiOCQA rewrite generation. Our approach can train effective rewrite models with a limited number of rewrite instances, without the need to derive the optimal labels.
% ince our goal is not to derive the best rewrite labels, but using a limited number of rewrite instances can help us to 

\begin{table}
\small
    \centering
    \colorbox{purple!8}{
    \begin{tabular}{@{}p{7.2cm}}
Given a question and its context, decontextualize the question by addressing coreference and omission issues. The resulting question should retain its original meaning and be as informative as possible, and should not duplicate any previously asked questions in the context.
\\ \\
Context: [Q: When was Born to Fly released? A: Sara Evans's third studio album, Born to Fly, was released on October 10, 2000.]\\Question: Was Born to Fly well received by critics?\\Rewrite: Was Born to Fly well received by critics?
\\ \\
Context: [Q: When was Keith Carradine born? A: Keith Ian Carradine was born August 8, 1949. Q: Is he married? A: Keith Carradine married Sandra Will on February 6, 1982.]\\Question: Do they have any children?\\Rewrite: Do Keith Carradine and Sandra Will have any children?
\\ \\
Context: [Q: Who proposed that atoms are the basic units of matter? A: John Dalton proposed that each chemical element is composed of atoms of a single, unique type, and they can combine to form more complex structures called chemical compounds.]\\Question: How did the proposal come about?\\Rewrite: How did John Dalton's proposal that each chemical element is composed of atoms of a single unique type, and they can combine to form more complex structures called chemical compounds come about?
\\ \\
Context: [Q: What is it called when two liquids separate? A: Decantation is a process for the separation of mixtures of immiscible liquids or of a liquid and a solid mixture such as a suspension. Q: How does the separation occur? A: The layer closer to the top of the container-the less dense of the two liquids, or the liquid from which the precipitate or sediment has settled out-is poured off.]\\Question: Then what happens?\\Rewrite: Then what happens after the layer closer to the top of the container is poured off with decantation?
\\ \\
Context: \{\texttt{current\_context}\}
\\
Question: \{\texttt{current\_question}\}
\\
Rewrite: 
    \end{tabular}
    }
    \caption{Prompt for rewrite generation. \citet{ye-etal-2023-enhancing} used this prompt to generate rewrite labels for QReCC under few-shot learning setting with ChatGPT (\texttt{gpt-3.5-turbo}). We follow the same prompt to derive rewrite labels for TopiOCQA. }
    \label{tab: rewrite-label-prompt}
\end{table}

\begin{table}
\small
\centering
\colorbox{green!6}{
\begin{tabular}{@{}p{7.2cm}}
Given a question and its context, decontextualize the question by addressing coreference and omission issues. The resulting question should retain its original meaning and be as informative as possible, and should not duplicate any previously asked questions in the context.
\\ \\
Context: \{\texttt{current\_context}\}
\\
Question: \{\texttt{current\_question}\}
\\
Rewrite: 
\end{tabular}
}

\caption{Prompt for GPT4o-0shot prompting on Doc2Dial and MultiDoc2Dial benchmarks. We use the same instruction as the rewrite generation in Table \ref{tab: rewrite-label-prompt}.}
\label{tab: gpt4o-prompt-0shot}
\end{table}

\begin{table}
\small
\centering
\colorbox{green!10}{
\begin{tabular}{@{}p{7.2cm}}
Given a question and its context, decontextualize the question by addressing coreference and omission issues. The resulting question should retain its original meaning and be as informative as possible, and should not duplicate any previously asked questions in the context.
\\ \\
Context: Q: When was Keith Carradine born? A: Keith Ian Carradine was born August 8, 1949. \\
Q: Is he married? A: Keith Carradine married Sandra Will on February 6, 1982.\\
Question: Do they have any children?\\
Rewrite: Do Keith Carradine and Sandra Will have any children?\\
\\ \\
Context: \{\texttt{current\_context}\}
\\
Question: \{\texttt{current\_question}\}
\\
Rewrite: 
    \end{tabular}
}

\caption{Prompt for GPT4o-1shot prompting on Doc2Dial and MultiDoc2Dial benchmarks. We use the same instruction as the rewrite generation in Table \ref{tab: rewrite-label-prompt}. The out-of-domain demonstration example is also used in Table \ref{tab: rewrite-label-prompt} by \citet{ye-etal-2023-enhancing}.}
\label{tab: gpt4o-prompt-1shot}
\end{table}

%% file: appendix-concate.tex
% \usepackage{multirow}
\begin{table}[ht]
\centering
\scalebox{0.7}{
\begin{tabular}{lccccccc}
\toprule
\multicolumn{8}{c}{\textbf{QReCC (8209)}} \\
 & MRR & MAP & NDCG & R@1 & R@5 & R@50 & avg \\ \midrule 
C5 & 51.9 & 50.3 & 49.5 & 38.9 & \textbf{61.4} & \multicolumn{1}{c|}{\textbf{85.7}} & 56.3 \\
M5 & \textbf{52.3} & \textbf{50.8} & \textbf{49.9} & \textbf{39.8} & 61.3 & \multicolumn{1}{c|}{85.0} & \textbf{56.5} \\
$\uparrow$ (\%) & 0.8 & 0.9 & 0.9 & 2.1 & -0.1 & \multicolumn{1}{c|}{-0.8} & 0.4 \\ \midrule
C9 & 51.0 & 49.4 & 48.5 & 37.8 & 60.8 & \multicolumn{1}{c|}{\textbf{86.1}} & 55.6 \\
M9 & \textbf{52.0} & \textbf{50.5} & \textbf{49.6} & \textbf{39.5} & \textbf{61.0} & \multicolumn{1}{c|}{84.7} & \textbf{56.2} \\
$\uparrow$ (\%)  & 2.1 & 2.2 & 2.3 & 4.5 & 0.3 & \multicolumn{1}{c|}{-1.6} & 1.1 \\ \midrule \midrule
\multicolumn{8}{c}{\textbf{Doc2Dial (640)}} \\
 & MRR & NDCG & R@1 & R@5 & R@10 & R@20 & avg \\ \midrule 
C5 & 59.1 & 58.6 & 45.5 & 75.9 & 83.1 & \multicolumn{1}{c|}{\textbf{91.4}} & 68.9 \\
M5 & \textbf{59.9} & \textbf{59.7} & \textbf{46.1} & \textbf{77.3} & \textbf{84.5} & \multicolumn{1}{c|}{90.0} & \textbf{69.6} \\
$\uparrow$ (\%)  & 1.3 & 1.8 & 1.4 & 1.8 & 1.7 & \multicolumn{1}{c|}{-1.5} & 0.9 \\ \midrule
C9 & 59.7 & 59.2 & 46.1 & 76.1 & 82.8 & \multicolumn{1}{c|}{\textbf{91.7}} & 69.3 \\
M9 & \textbf{60.3} & \textbf{60.1} & \textbf{46.7} & \textbf{77.0} & \textbf{85.6} & \multicolumn{1}{c|}{90.9} & \textbf{70.1} \\
$\uparrow$ (\%)  & 1.1 & 1.6 & 1.4 & 1.2 & 3.4 & \multicolumn{1}{c|}{-0.9} & 1.2 \\ \bottomrule
\end{tabular}}
\caption{Comparison between two passage organization types in reward calculation with \texttt{QReCC-SFT} setting: \texttt{Concatenated (C)} vs.  \texttt{Marginalized (M)}. We consider two $K$ values, top-5 passages (\texttt{C5} and \texttt{M5}) and top-9 passages (\texttt{C9} and \texttt{M9}) since \texttt{C1} and \texttt{M1} are identical. $\uparrow$ (\%) denotes the percentage improvement of \texttt{M} over \texttt{C}, relative to \texttt{C}.}
\label{tab: appendix-concate}
\end{table}

%% file: algorithm.tex
\begin{algorithm*}[t]
\caption{AdaQR Algorithm}
    \label{algo}
    \begin{algorithmic}[1]
    \REQUIRE LLM $\mathcal{M_{\theta}}$, Pre-trained LLM $\mathcal{A}$, retriever $\mathcal{R}$, seed dataset $D_S=\{H, q, r\}$, target dataset $D_T=\{H, q, a\}$ with corpus $P$, threshold $\delta$
    \ENSURE Aligned query rewriting LLM $\mathcal{M_{\theta}}$
    \STATE \colorbox[HTML]{ECF4FF}{// Supervised Fine-Tuning}
    \STATE initialize $\mathcal{M}_{SFT}=\mathcal{M_{\theta}}$ and fine-tune $\mathcal{M}_{SFT}$ on $D_S$ using $\mathcal{L}_{SFT} = -\log p_{\mathcal{M_{\theta}}}(r|H, q)$
    % \STATE $\mathcal{M_{SFT}} = \min_{(H_{<t}, q_t, r^*_t)\in D_S} -\log p_{\mathcal{M}}(r_t|H_{<t}, q_t)$
    % \STATE $\mathcal{M_{SFT}} = \min_{(H_{<t}, q_t, r^*_t)\in D_S} -\log p_{\mathcal{M}}(r_t|H_{<t}, q_t)$
    \FOR{$(H_{<t}, q_t, a_t) \in D_{T}$}
        \STATE $\hat{r}^1, \hat{r}^2, \hat{r}^3 \sim \mathcal{M_{SFT}}(\cdot |H, q)$ $\triangleright$ sample $3$ rewrite candidates from $\mathcal{M_{SFT}}$
        
        \STATE \colorbox[HTML]{ECF4FF}{// Reward Collection}
        \FOR{$i \in \{1,2,3\}$}
            \STATE $e^i, \mathcal{X} \leftarrow 0, \emptyset$
            \STATE $P^i=\{p^{i}_1, ..., p^{i}_k\}_{k=1}^K \sim \mathcal{R}(\hat{r}^i, P)$ $\triangleright$ retrieve top-$K$ passages for each rewrite with retriever $\mathcal{R}$
            \FOR{$k \in \{1,...,K\}$}
                \STATE 
                % $\mathcal{L}_{CQA}^k=-\log p_{\mathcal{A}}(a_t|H_{<t}, q_t,p^k)$ 
                $\mathcal{S}_k=\log p_{\mathcal{A}}(a|H, q,p^i_{k})$
                $\triangleright$ compute the log probability of answer conditioned on the retrieved passage and conversational query
                \STATE $e^i \mathrel{+}= \mathcal{P}_{\mathcal{R}}(p_k^i|\hat{r}^i)\mathcal{S}_k$
            \ENDFOR
        \ENDFOR
    
        \STATE \colorbox[HTML]{ECF4FF}{// Preference Pairs Construction}
        \FOR{$w \in \{1,2,3\}$}
            \FOR{$l \in \{1,2,3\}$ and $l > w$}
                \IF{$e^w - e^l > \delta$}
                    \STATE $\mathcal{X} \leftarrow \mathcal{X} \cup {(H, q, r^w, r^l)}$
                \ENDIF
            \ENDFOR
        \ENDFOR
    \ENDFOR
    \STATE \colorbox[HTML]{ECF4FF}{// Preference Optimization}
    \STATE initialize $\mathcal{M}_{\theta}=\mathcal{M}_{SFT}$ and fine-tune $\mathcal{M}_{\theta}$ on $\mathcal{X}$ using \\

    $\mathcal{L}_{DPO}=-\log \sigma(\beta \log\frac{\mathcal{M}_{\theta}\left(r_{w} \mid q_{t}, H_{<t}\right)}{\mathcal{M}_{SFT}\left(r_{w} \mid q_{t}, H_{<t}\right)}-\beta \log \frac{\mathcal{M}_{\theta}\left(r_{l} \mid q_{t}, H_{<t}\right)}{\mathcal{M}_{SFT}\left(r_{l} \mid q_{t}, H_{<t}\right)})$
    \RETURN $\mathcal{M}_{\theta}$
    \end{algorithmic}
    \label{algorithm}
\end{algorithm*}

%% file: appendix-data.tex
\begin{table*}[ht]
\centering
\scalebox{0.8}{
\begin{tabular}{l|ccc}
\toprule
\textbf{Dataset} & \textbf{Train (all/w-ans/w-ans+non-turn1)} & \textbf{Evaluation (all/w-psg/w-psg+non-turn1)} & \textbf{Corpus Size} \\ \midrule
QReCC & 51928/47463/39677 & 16451/8209/6852 & $\sim$50M \\
TopiOCQA & 45450/41798/38862 & 2514/2514/2309 & $\sim$50M \\
Doc2Dial & 21998/21998/10011 & 640/640/640 & 1557 \\
MultiDoc2Dial & 24603/24603/11101 & 648/648/648 & 1559 \\ \bottomrule
\end{tabular}}
\caption{Data statistics of QReCC, TopiOCQA, Doc2Dial and MultiDoc2Dial. For training split, we report the total number of instances (\texttt{all}), the number of instances with answers (\texttt{w-ans}), and the number of instances with answers that are not first turns (\texttt{w-ans+non-turn1}). For evaluation, we report the total number of instances (\texttt{all}), the number of instances with passage labels (\texttt{w-psg}), and the number of instances with passage labels that are not first turns (\texttt{w-psg+non-turn1}) since reference labels are needed for metric scores calculation.}
\label{tab: appendix-data}
\end{table*}

%% file: appendix-gemma-llama.tex
\begin{table*}[ht]
\centering
\scalebox{0.68}{
\begin{tabular}{llcccccc|cccccc} 
\toprule
 & \multicolumn{7}{c|}{Gemma-7B} & \multicolumn{6}{c}{Llama-7B} \\ \midrule \midrule
\multicolumn{1}{l|}{} & \textbf{Method} & \textbf{MRR} & \textbf{MAP} & \textbf{NDCG} & \textbf{R@1} & \textbf{R@5} & \textbf{R@50} & \textbf{MRR} & \textbf{MAP} & \textbf{NDCG} & \textbf{R@1} & \textbf{R@5} & \textbf{R@50} \\ \cline{2-14} 
\multicolumn{1}{l|}{} & QReCC-SFT & 45.7 & 44.2 & 42.9 & 32.3 & 55.5 & 83.6 & 45.4 & 43.9 & 42.4&32.1&55.3 & 83.4 \\
\multicolumn{1}{l|}{} & + Pseudo-Label & 50.7 & 49.1 & 48.0 & 37.5 & 60.3 & 86.3 & 50.9&49.3&48.3	&37.8&60.7&85.2 \\
\multicolumn{1}{l|}{} & \cellcolor[HTML]{ECF4FF}+ Gold-Label & \cellcolor[HTML]{ECF4FF}51.2 & \cellcolor[HTML]{ECF4FF}49.7 & \cellcolor[HTML]{ECF4FF}48.7 & \cellcolor[HTML]{ECF4FF}37.8 & \cellcolor[HTML]{ECF4FF}61.3 & \cellcolor[HTML]{ECF4FF}\textbf{86.7} & \cellcolor[HTML]{ECF4FF}51.3 & \cellcolor[HTML]{ECF4FF}49.7 & \cellcolor[HTML]{ECF4FF}48.7& \cellcolor[HTML]{ECF4FF}38.2 & \cellcolor[HTML]{ECF4FF}\textbf{60.9} & \cellcolor[HTML]{ECF4FF}\textbf{85.7} \\
\multicolumn{1}{l|}{\multirow{-5}{*}{QReCC}} & \textbf{+ Ours} & \textbf{52.5} & \textbf{50.8} & \textbf{50.1} & \textbf{39.4} & \textbf{61.7} & 84.8 & \textbf{52.0}&\textbf{50.4}&\textbf{49.6} & \textbf{39.5}&\textbf{60.9}&83.8\\ \midrule \midrule
\multicolumn{1}{l|}{} & \textbf{Method} & \textbf{MRR} & \textbf{NDCG} & \textbf{R@1} & \textbf{R@5} & \textbf{R@50} & \textbf{R@100} & \textbf{MRR} & \textbf{NDCG} & \textbf{R@1} & \textbf{R@5} & \textbf{R@50} & \textbf{R@100} \\ \cline{2-14} 
\multicolumn{1}{l|}{} & QReCC-SFT & 16.4 & 14.7 & 10.5 & 22.6 & 45.4 & 52.7 & 15.3 & 13.6 & 9.2 & 21.3 & 45.0 & 51.4 \\
\multicolumn{1}{l|}{} & + Pseudo-Label & 13.4 & 11.8 & 8.6 & 18.0 & 38.4 & 45.7 & 12.87 & 11.4 & 8.4 & 16.9 & 37.4 & 45.1 \\
\multicolumn{1}{l|}{} & \cellcolor[HTML]{ECF4FF} + Gold-Label & \cellcolor[HTML]{ECF4FF}\textbf{20.5} & \cellcolor[HTML]{ECF4FF}18.3 & \cellcolor[HTML]{ECF4FF}12.5 & \cellcolor[HTML]{ECF4FF}\textbf{28.6} & \cellcolor[HTML]{ECF4FF}\textbf{57.8} & \cellcolor[HTML]{ECF4FF}\textbf{65.8} & \cellcolor[HTML]{ECF4FF}\textbf{19.4} & \cellcolor[HTML]{ECF4FF}\textbf{17.2} & \cellcolor[HTML]{ECF4FF}11.9 & \cellcolor[HTML]{ECF4FF}\textbf{27.0} & \cellcolor[HTML]{ECF4FF}\textbf{56.2} & \cellcolor[HTML]{ECF4FF}\textbf{64.5} \\
\multicolumn{1}{l|}{\multirow{-5}{*}{TopiOCQA}} & \textbf{+ Ours} & 20.4 & \textbf{18.5} & \textbf{13.1} & 28.2 & 54.8 & 62.7 & 18.86 & \textbf{17.2} & \textbf{12.3} & 25.5 & 50.8 & 58.8 \\ \midrule \midrule
\multicolumn{1}{l|}{} & \textbf{Method} & \textbf{MRR} & \textbf{NDCG} & \textbf{R@1} & \textbf{R@5} & \textbf{R@10} & \textbf{R@20} & \textbf{MRR} & \textbf{NDCG} & \textbf{R@1} & \textbf{R@5} & \textbf{R@10} & \textbf{R@20} \\ \cline{2-14} 
\multicolumn{1}{l|}{} & QReCC-SFT & 57.9 & 57.2 & 44.7 & 75.0 & 82.5 & 90.8 & 56.2 & 55.8 & 42.3 & 74.1 & 80.9 & 88.3 \\
\multicolumn{1}{l|}{} & + Pseudo-Label & 58.8 & 58.5 & 44.7 & 76.3 & 84.2 & 91.6 & 58.9 & 59.1 & 44.5 & 75.8 & 84.1 & 90.5 \\
\multicolumn{1}{l|}{} & \cellcolor[HTML]{ECF4FF}+ Gold-Label & \cellcolor[HTML]{ECF4FF}60.3 & \cellcolor[HTML]{ECF4FF}60.2 & \cellcolor[HTML]{ECF4FF}46.1 & \cellcolor[HTML]{ECF4FF}78.3 & \cellcolor[HTML]{ECF4FF}85.6 & \cellcolor[HTML]{ECF4FF}93.0 & \cellcolor[HTML]{ECF4FF}\textbf{60.3} & \cellcolor[HTML]{ECF4FF}\textbf{60.2} & \cellcolor[HTML]{ECF4FF}\textbf{46.6} & \cellcolor[HTML]{ECF4FF}\textbf{75.9} & \cellcolor[HTML]{ECF4FF}\textbf{85.3} & \cellcolor[HTML]{ECF4FF}91.6 \\
\multicolumn{1}{l|}{\multirow{-5}{*}{Doc2Dial}} & \textbf{+ Ours} & \textbf{60.9} & \textbf{60.8} & \textbf{46.9} & \textbf{78.8} & \textbf{86.1} & \textbf{93.1} & 59.3 & 59.5 & 45.6 & 74.7 & 84.5 & \textbf{91.7} \\ \midrule \midrule
\multicolumn{1}{l|}{} & \textbf{Method} & \textbf{MRR} & \textbf{NDCG} & \textbf{R@1} & \textbf{R@5} & \textbf{R@10} & \textbf{R@20} & \textbf{MRR} & \textbf{NDCG} & \textbf{R@1} & \textbf{R@5} & \textbf{R@10} & \textbf{R@20} \\ \cline{2-14} 
\multicolumn{1}{l|}{} & QReCC-SFT & 52.0 & 50.4 & 40.1 & 65.4 & 74.7 & 84.3 & 51.2 & 50.8 & 37.8 & 67.0 & 75.5 & 83.2 \\
\multicolumn{1}{l|}{} & + Pseudo-Label & 53.3 & 52.0 & 41.1 & 67.4 & 76.1 & 86.1 & 54.7 & 54.3 & 41.7 & 69.9 & 78.1 & 86.0 \\
\multicolumn{1}{l|}{} & \cellcolor[HTML]{ECF4FF}+ Gold-Label & \cellcolor[HTML]{ECF4FF}54.7 & \cellcolor[HTML]{ECF4FF}53.7 & \cellcolor[HTML]{ECF4FF}42.0 & \cellcolor[HTML]{ECF4FF}68.4 & \cellcolor[HTML]{ECF4FF}79.0 & \cellcolor[HTML]{ECF4FF}\textbf{88.1} & \cellcolor[HTML]{ECF4FF}55.2 & \cellcolor[HTML]{ECF4FF}55.0 & \cellcolor[HTML]{ECF4FF}41.5 & \cellcolor[HTML]{ECF4FF}71.0 & \cellcolor[HTML]{ECF4FF}78.4 & \cellcolor[HTML]{ECF4FF}87.2 \\
\multicolumn{1}{l|}{\multirow{-5}{*}{MultiDoc2Dial}} & \textbf{+ Ours} & \textbf{55.1} & \textbf{53.9} & \textbf{42.1} & \textbf{70.2} & \textbf{80.1} & \textbf{88.1} & \textbf{56.5} & \textbf{55.8} & \textbf{43.4} & \textbf{73.3} & \textbf{80.4} & \textbf{87.8} \\ \bottomrule
\end{tabular}}
\caption{Evaluation results of sparse retrieval (BM25) on QReCC, TopiOCQA, Doc2Dial and MultiDoc2Dial with Gemma-7b and Llama-7b. Due to computational constraints, we experiment AdaQR with QReCC as seed dataset only.}
\label{tab:gemma-llama}
\end{table*}

%% file: analysis-top-k.tex
\begin{table*}[ht]
\centering
\scalebox{0.77}{
\begin{tabular}{c|ccccccc|cccccc}
\toprule
 & \multicolumn{7}{c|}{\textbf{QReCC (8209)}} & \multicolumn{6}{c}{\textbf{Doc2Dial (640)}} \\
 & \textbf{R@1} & \textbf{R@5} & \textbf{R@10} & \textbf{R@50} & \textbf{MRR} & \textbf{MAP} & \textbf{NDCG} & \textbf{R@1} & \textbf{R@5} & \textbf{R@10} & \textbf{R@20} & \textbf{MRR} & \textbf{NDCG} \\ \midrule \midrule
QReCC SFT & 32.4 & 56.0 & 64.9 & 83.7 & 45.9 & 44.4 & 43.2 & 42.3 & 72.0 & 81.7 & 89.5 & 56.0 & 55.7 \\ \midrule
\ + Top-1 & 39.4 & 61.0 & 68.6 & 84.4 & 52.0 & 50.4 & 49.6 & 45.3 & 76.1 & 82.7 & 90.0 & 59.2 & 59.2 \\
\ + Top-3 & 39.6 & 61.1 & 68.7 & 84.7 & 52.1 & 50.6 & 49.7 & 45.0 & 77.0 & 84.5 & \textbf{91.1} & 59.3 & 59.2 \\
\ + Top-5 & \textbf{39.8} & \textbf{61.3} & \textbf{69.1} & \textbf{85.0} & 52.3 & \textbf{50.8} & \textbf{49.9} & 46.1 & 77.3 & 84.5 & 90.0 & 59.9 & 59.7 \\
\ + Top-7 & 39.7&\textbf{61.3}&69.0&84.9&\textbf{52.4}&50.7&\textbf{49.9}  & 46.4 & \textbf{77.7} & 85.0 & 90.2 & 60.2 & \textbf{60.2} \\
\ + Top-9 &39.5&61.0&68.9&84.7&52.0&50.5&49.6& \textbf{46.7} & 77.0 & \textbf{85.6} & 90.9 & \textbf{60.3} & 60.1 \\ \bottomrule
\end{tabular}}
\caption{Evaluation results of sparse retrieval (BM25) on QReCC and Doc2Dial with varying top-$K$ values ($K=1,3,5,7,9$) during reward calculation with \texttt{QReCC-SFT} setting.}
\label{tab: analysis-topk}
\end{table*}

%% file: appendix-data-volume.tex
\begin{table*}[ht]
\centering
\scalebox{0.68}{
\begin{tabular}{r|ccccccccc|ccccccc}
\toprule
 & \multicolumn{9}{c|}{\textbf{QReCC (8209)}} & \multicolumn{7}{c}{\textbf{Doc2Dial (640)}} \\
 & MRR & MAP & NDCG & R@1 & R@5 & R@10 & R20 & R@50 & AVG & MRR & NDCG & R@1 & R@5 & R@10 & R@20 & \multicolumn{1}{c}{AVG} \\ \midrule \midrule
QReCC SFT & 45.9 & 44.4 & 43.2 & 32.4 & 56.0 & 64.9 & 73.6 & \multicolumn{1}{c|}{83.7} & 55.5 & 56.0 & 55.7 & 42.3 & 72.0 & 81.7 & \multicolumn{1}{c|}{89.5} & 66.2 \\ \hline
+ 20\% Data & 51.2 & 49.6 & 48.7 & 38.5 & 60.5 & 68.8 & 76.2 & \multicolumn{1}{c|}{84.0} & 59.7 & 59.4 & 59.2 & 45.6 & 75.9 & 83.3 & \multicolumn{1}{c|}{\textbf{91.6}} & 69.2 \\
+ 40\% Data & 51.7 & 50.1 & 49.2 & 39.1 & 60.8 & 68.4 & 76.0& \multicolumn{1}{c|}{84.1} & 59.9 & 59.6 & 59.4 & 45.9 & 76.7 & 83.4 & \multicolumn{1}{c|}{90.9} & 69.3 \\
+ 60\% Data & 51.7 & 50.1 & 49.2 & 39.1 & 60.6 & 68.4 & 76.0& \multicolumn{1}{c|}{84.2} & 59.9 & 59.4 & 59.4 & 45.6 & 76.3 & 83.6 & \multicolumn{1}{c|}{90.6} & 69.1 \\
+ 80\% Data & 52.0 & 50.4 & 49.5 & 39.6 & 60.8 & 68.5 &76.2 & \multicolumn{1}{c|}{84.5} & 60.2 & 59.5 & \textbf{59.7} & 45.3 & \textbf{77.3} & \textbf{84.5} & \multicolumn{1}{c|}{90.6} & 69.5 \\
+ 100\% Data & \textbf{52.3} & \textbf{50.8} & \textbf{49.9} & \textbf{39.8} & \textbf{61.3} & \textbf{69.1} & \textbf{76.7} & \multicolumn{1}{c|}{\textbf{85.0}} & \textbf{60.6} & \textbf{59.9} & \textbf{59.7} & \textbf{46.1} & \textbf{77.3} & \textbf{84.5} & \multicolumn{1}{c|}{90.0} & \textbf{69.6} \\ \bottomrule
\end{tabular}}
\caption{Evaluation results of sparse retrieval (BM25) on QReCC and Doc2Dial with varying number of training data ($20\%$, $40\%$, $60\%$, $80\%$ and $100\%$) during preference alignment with \texttt{QReCC-SFT} setting.}
\label{tab: appendix-data-volume}
\end{table*}